\documentclass[twocloums]{IEEEtran}

\usepackage{cite}
\usepackage{CJKutf8}

\ifCLASSINFOpdf
 
\else
 
\fi
\usepackage{tikz}
\usepackage{pifont}
\usepackage{bbding}
\usetikzlibrary{matrix}
\usetikzlibrary{decorations.pathreplacing}
\usepackage{graphicx}
\usepackage[colorlinks,linkcolor=red]{hyperref}

\graphicspath{{images/}}
\usepackage{enumitem}
\usepackage{amsmath}
\usepackage{subfigure}
\usepackage{float}
\usepackage{cite}
\usepackage{svg}
\usepackage{amsmath}
\usepackage{graphics}
\usepackage{graphicx}
\usepackage{amssymb}
\usepackage[ruled,linesnumbered]{algorithm2e} 

\usepackage{mathrsfs}
\usepackage{siunitx}
\usepackage{booktabs}
\usepackage{bm}
\usepackage{bbm}

\usepackage{soul}
\usepackage{color, xcolor}  

\usepackage{float}
\usepackage{url}
\usepackage{amsfonts}
\usepackage[square,sort&compress,numbers]{natbib}
\usepackage{multirow}
\usepackage{textcomp}
\usepackage{supertabular}
\usepackage{array}
\usepackage{pifont}
\usepackage{hyperref}

\usepackage{soul}
\soulregister\cite7 
\soulregister\citep7 
\soulregister\citet7 
\soulregister\ref7 
\soulregister\pageref7 
\soulregister\section7
\soulregister\Function7
\soulregister\State7
\soulregister\EndFunction7
\soulregister\textbf7
\soulregister\Return7

\usepackage{algpseudocode}  
\usepackage{amsmath} 

\begin{document}

\title{Implicit Ray-Transformers for Multi-view Remote Sensing Image Segmentation}

\author{Zipeng Qi, Hao Chen, Chenyang Liu, Zhenwei Shi and Zhengxia Zou$^\star$,~\IEEEmembership{Member,~IEEE} 
\thanks{The work was supported by the National Key Research and Development Program of China (Grant No. 2022ZD0160401), the National Natural Science Foundation of China under Grant 62125102, and the Fundamental Research Funds for the Central Universities.
\emph{(Corresponding author: Zhengxia Zou (e-mail: zhengxiazou@buaa.edu.cn))}}
\thanks{Zipeng Qi, Hao Chen, Chenyang Liu and Zhenwei Shi are with the Image Processing Center, School of Astronautics, and with the Beijing Key Laboratory of Digital Media, and with the State Key Laboratory of Virtual Reality Technology and Systems, Beihang University, Beijing 100191, China, and also with the Shanghai Artificial Intelligence Laboratory, Shanghai 200232, China.}
\thanks{Zhengxia Zou is with the Department of Guidance, Navigation and Control, School of Astronautics, Beihang University, Beijing 100191, China, and also with Shanghai Artificial Intelligence Laboratory, Shanghai 200232, China.}
}

\maketitle

\begin{abstract}
The mainstream CNN-based remote sensing (RS) image semantic segmentation approaches typically rely on massive labeled training data. Such a paradigm struggles with the problem of RS multi-view scene segmentation with limited labeled views due to the lack of considering 3D information within the scene. 
In this paper, we propose ``Implicit Ray-Transformer (IRT)'' based on Implicit Neural Representation (INR), for RS scene semantic segmentation with sparse labels (such as 4-6 labels per 100 images). We explore a new way of introducing multi-view 3D structure priors to the task for accurate and view-consistent semantic segmentation. The proposed method includes a two-stage learning process. In the first stage, we optimize a neural field to encode the color and 3D structure of the remote sensing scene based on multi-view images. In the second stage, we design a Ray Transformer to leverage the relations between the neural field 3D features and 2D texture features for learning better semantic representations. Different from previous methods that only consider 3D prior or 2D features, we incorporate additional 2D texture information and 3D prior by broadcasting CNN features to different point features along the sampled ray. 
To verify the effectiveness of the proposed method, we construct a challenging dataset containing six synthetic sub-datasets collected from the Carla platform and three real sub-datasets from Google Maps. Experiments show that the proposed method outperforms the CNN-based methods and the state-of-the-art INR-based segmentation methods in quantitative and qualitative metrics. Ablation study shows that under limited label conditions, the combination of the 3D structure prior and 2D texture can significantly improve the performance and effectively complete missing semantic information in novel views. Experiments also demonstrate the proposed method could yield geometry-consistent segmentation results against illumination changes and viewpoint changes. Our data and code will be public.

\end{abstract}


\IEEEpeerreviewmaketitle

\section{Introduction}
\label{sec:introduction}


Remote sensing image segmentation is a fundamental yet challenging task that has been widely applied in various fields, such as cloud detection~\cite{li2020deep}, change detection~\cite{chen2021remote}, and land analysis~\cite{al2019land}. The objective of remote sensing image segmentation is to produce pixel-wise classified labels for an image. Thanks to the advancement of imaging technology and satellite networking technology, it is now easy to acquire high-resolution multi-view RGB images of remote sensing scenes. 

\begin{figure*}[t]
\centering
\includegraphics[width=\linewidth]{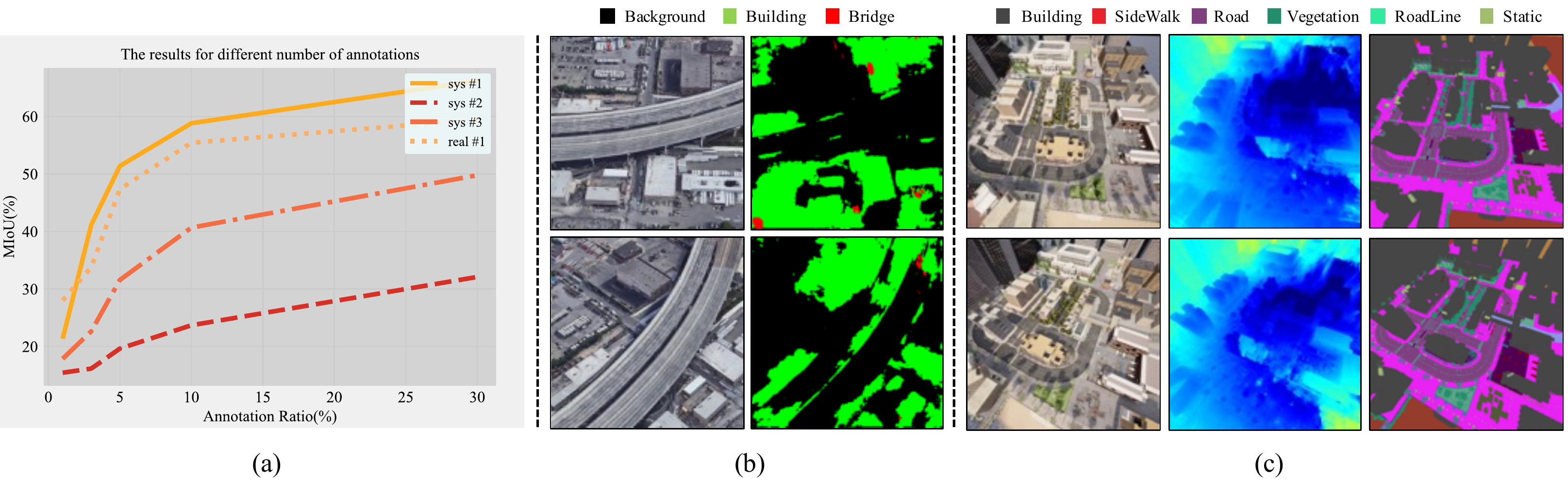}
\caption{(a): the performance of CNN-based methods with different annotation ratios; (b): it is difficult for the CNN-based methods to distinguish objects with similar textures under the sparse annotations; (c): the correspondence of 3D structures to objects and labels.}
\label{Fig1}
\end{figure*}

The mainstream segmentation methods ~\cite{ronneberger2015u, chen2017rethinking, fu2019dual} benefit from the deep convolution neural networks (CNN), which can effectively learn and extract robust and discriminative features from the input images. However, deep CNN-based remote sensing image segmentation methods rely heavily on massive training data. As shown in Fig.~\ref{Fig1}(a), the performance of traditional CNN-based methods is sensitive to the number of annotations. A large number of high-quality pixel-wise annotations, as a guarantee for the performance of CNN-based segmentation methods, consume a great deal of time and effort. As for the task of semantic segmentation for a 3D scene given only limited annotated views, the CNN-based methods may overfit the views in the training data but generate poor results for the rest of the views. The key reason is that the 2D texture information or 2D context relationship is insufficient to identify similar-textured objects (Fig. \ref{Fig1}(b)) in a 3D scene. Finally, the 3D context relationship of a scene is also crucial for semantic attribute prediction (Fig.\ref{Fig1}(c)). For example, the building is typically higher than the road and the same object across different views usually has a similar texture. However, these properties have been rarely investigated in previous papers.



\begin{figure}[t]
\centering
\includegraphics[width=\linewidth]{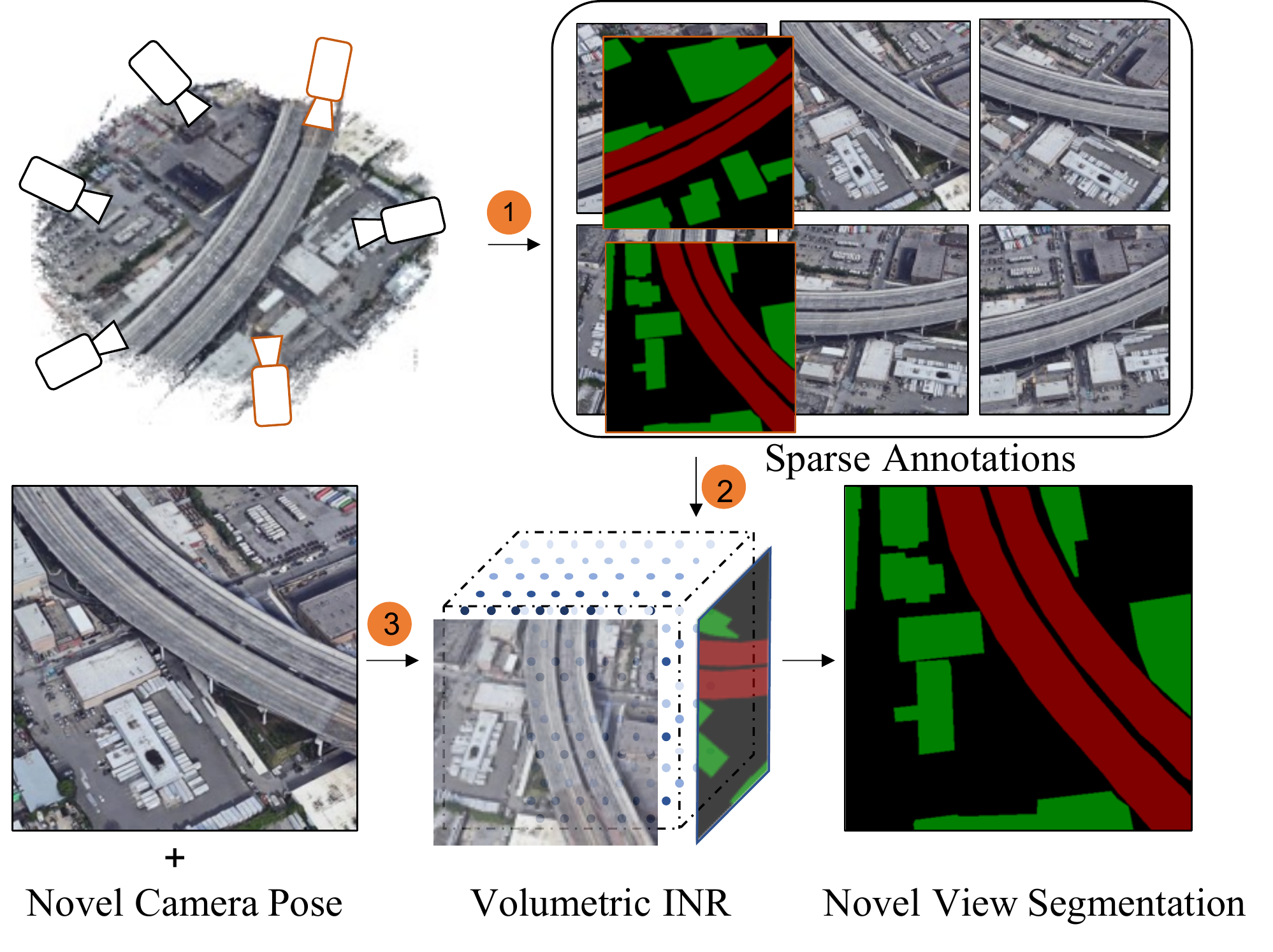}
\caption{In this paper, we propose an INR-based method to combine the 3D ray-semantic features and 2D CNN features for remote sensing scene semantic segmentation. \ding{172}: the multi-view images capturing and sparse annotations labeling; \ding{173}: the INR construction; \ding{174}: novel view segmentation generation.}
\label{problem}
\end{figure}

Considering the above challenges, this paper studies the task of multi-view remote sensing scene semantic segmentation under limited annotations, as shown in Fig. ~\ref{problem}. We show that the 3D structure prior is crucial for this task. The representation of the 3D structure is a fundamental and long-studied problem in computer vision and graphics. There are many explicit representation-based 3D reconstruction methods~\cite{pan2019deep, kanazawa2018learning, poullis2009automatic} that extract 3D context information in different forms, including depth map~\cite{ye2021shelf}, mesh~\cite{hu2021vmnet}, and point cloud~\cite{song2022lslpct}. However, these methods typically require explicit supervision data which are hard to obtain and computationally intensive. Recently, an emerging research topic named ``Implicit Neural Representation (INR)''  has made rapid advances, particularly in the realm of novel view synthesis~\cite{mildenhall2021nerf, zhang2020nerf++, barron2021mip, 9852475}. The INR provides a novel way to parameterize continuous signals driven by coordinates with neural networks. The INR-based novel view synthesis fits images from known viewpoints and utilizes the continuity of the spatially-varying scene properties (such as color and geometry) in high-dimensional space to render compelling photo-realistic novel view images. During this process, the geometry and color attributes are encoded into the weights of a neural network. Compared to explicit reconstruction, the INR is more fixable to optimized without expensive supervision.

In this paper, inspired by the INR in novel view synthesis, we propose a new framework for multi-view remote sensing image segmentation under limited view annotations which utilizes an INR to exploit 3D structure prior. We also proposed a new network architecture called ``Ray-Transformer'' that combines the 3D structure and 2D texture information from a set of 3D location points along the rays.
We refer to our method as ``Ray-Transformer'' and refer to the task we studied as Remote Sensing Scene Semantic Segmentation (R4S). Given a set of multi-view RGB images from a remote sensing scene, the proposed method can generate accurate and semantic-consistent novel view segmentation output even only trained with a limited number of labels (e.g., 4-6 labels per 100 images).

The proposed method has a two-stage learning process. We first optimize a color-INR of the target scene using multi-view RGB images, where the 3D context information is encoded in the weights of a set of MLPs. Then we employ a knowledge distillation strategy to convert color INR to semantic INR. In order to enhance the semantic consistency between multi-viewpoints. We design the Ray-Transformer to integrate and transfer the 3D ray-color features into ray-semantic features. Specially, we add a CNN texture token to broadcast texture information among different locations along a ray. Finally, we combine the 3D ray-semantic features from semantic-INR and 2D features from additional CNN to complete the missing semantic information in novel views and get more detail and accurate results.

Extensive experiments are conducted to verify the effectiveness of the method. We constructed six sets of synthesis data based on the well-known Carla simulation platform ~\cite{dosovitskiy2017carla}. We also construct three sets of real data from Google Maps. Our method outperforms CNN-based methods and INR-based state-of-the-art methods. The visual comparison also suggests that the proposed method can produce more accurate and visually consistent results. In addition, experiment results also show that our method has better performance against illumination and viewpoint changes. Our code and dataset will be made publicly available at \href{https://qizipeng.github.io/IRT}{https://qizipeng.github.io/IRT}.

The contributions of this paper are summarized as follows:

\begin{itemize}
	\item We propose a new method for multi-view remote sensing image segmentation method based on the implicit neural representation which combines the 2D CNN features with 3D ray-semantic features. Given a set of multi-view RGB images and a limited number of annotations, the proposed method effectively generates accurate segmentation results for novel views.
 	\item  We propose a density-driven and memory-friendly network architecture called Ray-Transfromer to integrate and transfer color features into semantic features; Specially we add the CNN texture token into the Ray-Transfromer and explore a different way of introducing texture information into the INR space.
	\item We constructed a challenging dataset for multi-view remote sensing image segmentation, which contains both real and synthetic images. Our method reached the state-of-the-art on the introduced dataset.
\end{itemize}

The rest of this paper is organized as follows. In Section \ref{section:related work}, we introduce the related work. In Section \ref{section:method}, we give a detailed introduction to the proposed method. In Section \ref{section:experiments}, the experimental results are presented. Conclusions are drawn in section \ref{section:conclusion}.

\section{Related Work}\label{section:related work}

\subsection{CNN-based Image Segmentation}

In the past few years, with the development of deep learning, Convolutional Neural Network (CNN)-based methods have become mainstream in image segmentation. The CNN-based methods typically employ an encoder-decoder structure.
There are mainly three groups of methods. 1) The first group adopts Unet~\cite{ronneberger2015u}-like architecture, where the skip connection is introduced to combine the low-level features into the decoder to keep more detailed information.  Some recent methods~\cite{yuan2021multi, jung2021boundary, li2021multistage, ghosh2018stacked} show the advantage of skip connection in remote sensing image segmentation tasks, \textit{e.g.} building detection, road detection, and multi-objects change detection.  2) The second group adopts a larger respective field and a deeper architecture to extract more semantic information. The dilated convolution was employed to enlarge the respective field of the networks, at the same to maintain a high output feature resolution~\cite{chen2017rethinking, chen2018encoder, hamaguchi2018effective, nogueira2019dynamic}. However, the stacking of multiple dilated convolutions may produce a gridding effect and thus a hybrid dilated convolution~\cite{wang2018understanding} is designed for alleviating this problem. 3) The third group employs a feature pyramid strategy to extract more contextual information from the remote sensing images~\cite{marmanis2018classification, wang2017gated, mou2020relation, zhou2020class}, especially for the images with multiple-scale objects. The above CNN-based methods have achieved good performance in many remote sensing segmentation problems, but they also have a common defect, that is, the training process usually relies on a large number of annotations. The main difference between our method and previous methods is that we use implicit neural representation, Transformer, and 3D scene structure prior to overcoming the problem of label dependence.

\subsection{Implicit Neural Segmentation}

Implicit neural representation is an emerging technique for characterizing continuous signals based on neural networks. In computer graphics and computer vision, implicit neural representations were first used to represent 3D scenes~\cite{mildenhall2021nerf,murez2020atlas, sun2021neuralrecon, qi20223d}. 
To optimize the neural network, these methods usually need 3D data for supervision. In the past two years, there are a variety of methods that optimized a volumetric space from a set of posed 2D images and do not need extra 3D supervision. NeRF~\cite{mildenhall2021nerf} is representative of these methods. Recently, implicit neural representations have also been introduced to image segmentation tasks~\cite{murez2020atlas,zhi2021place,fu2022panoptic, kundu2022panoptic,qi2022remote}, including indoor-scene image segmentation~\cite{zhi2021place}, traffic-scene image segmentation~\cite{fu2022panoptic, kundu2022panoptic} and remote sensing scene image segmentation~\cite{qi2022remote}. These methods can be divided into two groups. The first kind~\cite{zhi2021place, tseng2022cla} considers the segmentation and novel view generation jointly and trains a multi-task representation where semantic features and color features share the same feature extractor but use different prediction heads. The second kind~\cite{eteke2022semantic, vora2021nesf, qi2022remote} first optimizes a color implicit neural representation of a scene and then transfers the color features into semantic features by fine-tuning or distillation. The color information can be more effectively introduced into INR space with multi-task representation. In addition to the above two groups of methods, there are also some methods~\cite{fu2022panoptic, tschernezki2022neural} that use extra 3D data or memory-intensive 3D convolution to improve the scene segmentation accuracy under a large number of annotations. The difference between all the above methods and our method is that we not only employ 3D information from the INR space but also combine the CNN features with a newly designed transformer, thus achieving accurate remote sensing scene image segmentation with limited annotations.


\subsection{Transformers}

The transformer was first introduced in 2017~\cite{vaswani2017attention} and has been widely used in NLP tasks~\cite{devlin2018bert, brown2020language, fedus2021switch}, which effectively solves the problem of long-range dependencies. Recently, the transformer architecture has been introduced to computer vision and remote sensing to extract global or long-range context features and shows comparable or even better performance than the CNN-based methods in various visual tasks, including image classification~\cite{dosovitskiy2020image}, object detection~\cite{carion2020end} and multimode tasks~\cite{radford2021learning, hu2021unit, chen2021remote}. The main two components in a transformer are encoders and decoders. The transformer encoders are to explore the multi-head attention modes of input. And the transformer decoders perform the cross-attention between encoder features and additional masked input to obtain the final results. In some vision tasks, there are also some approaches that only use the transformer encoder and combined it with a CNN-based decoder. SETR~\cite{zheng2021rethinking} utilizes the transformer-based backbone and a standard CNN decoder to achieve image segmentation without decreasing the feature map resolution. Swin-transformer~\cite{liu2021swin} uses a variant of ViT~\cite{dosovitskiy2020image} composed of local shifting windows and a pyramid FCN decoder, which achieves the state-of-the-art performance in classification and segmentation tasks.
The transformer-based models usually come with larger computations and a more complex training pipeline. In our work, we design a memory-friendly transformer that works in ray space to only consider the valid 3D point features.

\begin{figure*}[t]
\centering
\includegraphics[width=\linewidth]{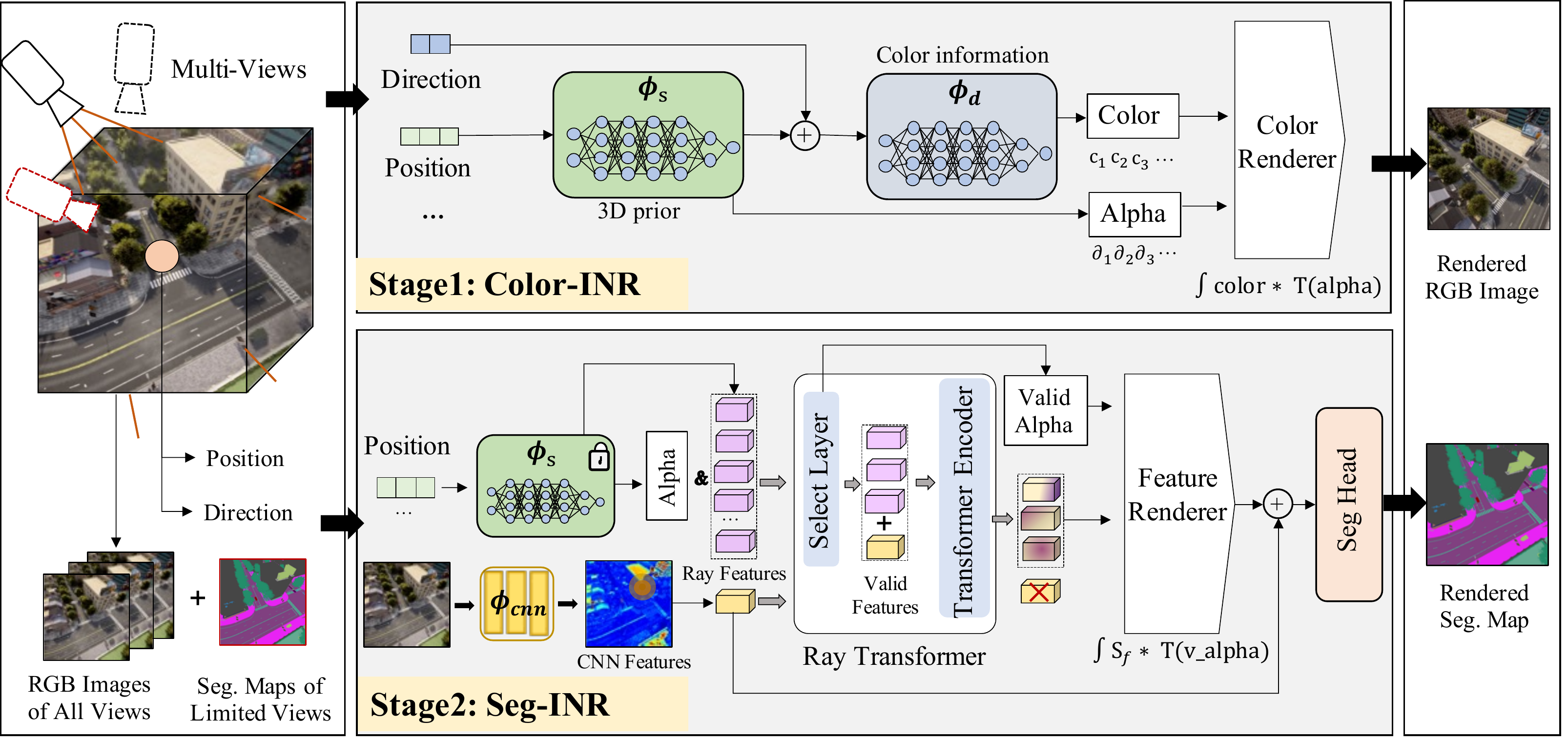}
\caption{The overview of the proposed model. The Color-INR take all the posed image as input and optimize the $\Phi_{s}$ and $\Phi_{d}$ to implicitly represent the scene. The 3D structure prior is encoded into the point features. The Seg-INR uses our designed Ray-Transformer to integrate and then covert the point features into ray-semantic features. In order to fully use the texture features in RGB images, we add a CNN token into the Ray-Transformer to complete information from unseen viewpoints. In final, we can achieve the novel view segmentation under the sparse labels for R4S task.}
\label{method}
\end{figure*}

\section{Methodology}\label{section:method}

An overview of our method is illustrated in Fig. \ref{method}. We propose an \textbf{I}mplicit \textbf{R}ay-\textbf{T}ransformer (\textbf{IRT}) for remote sensing scene semantic segmentation. The input of IRT is a group of posed images and a limited number of pixel-wise annotations. The output is the segmentation label maps of any novel views. It includes a two-stage learning process. In the first stage, we optimize a color-INR model using all posed images. The input of $\Phi_{s}$ is the position and view direction of sampled points along a random ray from the voxelized space of the scene. The output is the color and density attribute of each point. Then we use volume rendering to render the final pixel color of the corresponding ray. In the second stage, we distillate the point features from $\Phi_{s}$ and convert the color features into semantic features using a memory-friendly transformer named Ray-Tranformer. In the Ray-Transformer, we take the additional CNN features as an additional texture token to broadcast the texture information into other point features. Similarly, we render the semantic attribute of all the points along a ray and get the final ray-semantic features of the corresponding ray. In order to further complete the missing semantic information under the sparse annotations, we combine the ray-semantic features with CNN features to get the final pixel class prediction. The inference detail of IRT is shown in the Algorithm \ref{algorthim_IRT}.

\begin{algorithm}[t] 
\caption{Implicit Ray-Transformer for R4S}  
\KwIn{$\{(\mathbf{I}_n, \mathbf{C}_n, \mathbf{L}_m)|n = 1 : N, m = 1:M\}$ (N images($I$) with camera parameters($C$) and M labels ($L$))}
\KwIn{$\mathbf{Iteration_1}$ (iterate number in step1)}
\KwIn{$\mathbf{Iteration_2}$ (iterate number in step2)}
\KwOut{IRT} 
\KwOut{$\mathbf{S}^{tgt}$ (novel view segmentation)}
\BlankLine
//training\\
\quad // step1: Color-INR construction\\
\For{$i$ in $1:\mathbf{Iteration_1}$}
{
// select a training image from $I_n$\\
$\mathrm{I_i},\mathrm{C_i} = \mathrm{Sample}(\mathrm{I_n}, \mathrm{C_n})$\\
// select B training points in $I_i$ with $C_i$\\
$\mathrm{(x,y,z,\alpha,\beta)^{B}} = \mathrm{SamplePoints}(\mathrm{I_i}, \mathrm{C_i})$\\
//render color results\\
$\mathrm{color^B} = \mathrm{R}(\Phi_{d}(\Phi_{s}\mathrm{(x,y,z)^B},\mathrm{(\alpha,\beta)^B)})$\\
//gradient descent to optimize Color-INR
}
\BlankLine
\quad //step2: Seg-INR construction\\
\For{$j$ in $1:\mathbf{Iteration_2}$}
{
// select a training image from $L_n$\\
$\mathrm{L_j}, \mathrm{C_j} = \mathrm{Sample}(\mathrm{L_m}, \mathrm{C_n})$\\
// select B training points in $I_j$ with $C_j$\\
$\mathrm{(x,y,z)^{B}} = \mathrm{SamplePoints}(\mathrm{I_j}, \mathrm{C_j})$\\
//render semantic results \\
//($\Phi_{R}$:Ray-Transformer, $\Phi_{C}$:CNN)\\
$\mathrm{color^B} = \mathrm{R}(\Phi_{R}(\Phi_{s}\mathrm{(x,y,z)^B},\Phi_{C}^B))$\\
//gradient descent to optimize Seg-INR
}
\BlankLine
//inference\\
\quad //render segmentation result from novel view\\
$\mathbf{S}^{tgt} = \mathbf{IRT}(\mathrm{(x,y,z)^{all}})$
\label{algorthim_IRT}  
\end{algorithm}

\subsection{Coordinate System Conversion}

In our approach, an important and fundamental operation is to sample points along a ray, which involves coordinate systems and their associated transformations. Thus, in this subsection, we first introduce the conversion between different coordinate systems, including the pixel coordinate system, image coordinate system, camera coordinate system, and world coordinate system.

The pixels coordinate system represents the projection of a 3D space object on the image plane. The coordinate origin is in the upper left corner of the CCD image plane. We take $(u,v)$ to represent the pixel coordinate. The origin of the image coordinate system is in the center of the CCD image plane, and its horizontal and vertical axes are parallel to the pixel coordinate system. We take $(x, y)$ to represent the image coordinate. The conversion relationship between the pixel coordinate system and the image coordinate system is defined as follows:
\begin{equation}
    \begin{bmatrix}
       u \\ v \\ 1 
   \end{bmatrix} = 
  \begin{bmatrix}
  \frac{1}{d_x}&0&u_0\\
  0&\frac{1}{d_y}&v_0\\
  0&0&1
  \end{bmatrix}
   \begin{bmatrix}
       x \\ y \\ 1 
   \end{bmatrix},
\end{equation}
where $d_x$ and $d_y$ respectively represent the width and height of a pixel corresponding to the photosensitive point. 
The camera coordinate system takes the optical center of the camera as the origin of the coordinate system. $x_c$, $y_c$ axes are parallel to the x and y axes of the image coordinate system. The optical axis of the camera is set to the $z_c$ axis and the coordinate system follows the right-hand rule.

The camera coordinate system to the image coordinate system follows a perspective transformation relationship, which can be represented by using similar triangles:
\begin{equation}
    z_c    
    \begin{bmatrix}
       x \\ y \\ 1 
   \end{bmatrix} =
   \begin{bmatrix}
  f&0&0&0\\
  0&f&0&0\\
  0&0&f&0\\
  0&0&1&0
  \end{bmatrix}
   \begin{bmatrix}
       x_c \\ y_c \\ z_c\\1 
   \end{bmatrix} =
  \begin{bmatrix}
    K|0  
  \end{bmatrix}
   \begin{bmatrix}
       x_c \\ y_c \\ z_c\\1 
   \end{bmatrix},
\end{equation}
where $f$ is the focal length and $K$ is the intrinsic parameter matrix of the camera. The world coordinate system can obtain the camera coordinate system through rotation and translation:
\begin{equation}
    \begin{bmatrix}
       x_c \\ y_c \\ z_c\\1 
   \end{bmatrix} =
       \begin{bmatrix}
       R & t \\ 
       0_{1\times 3}&1 
   \end{bmatrix}
   \begin{bmatrix}
       x_w \\ y_w \\ z_w\\1 
   \end{bmatrix},
\end{equation}
where $R$ is the rotation matrix and $t$ is the translation matrix. $[R|t]$ is the extrinsic parameter of the camera. $x_w$, $y_w$, and $z_w$ are the absolute coordinates of space objects in the world coordinate system. In our work, we use the COLMAP~\cite{schonberger2016structure} to estimate the intrinsic and extrinsic parameters of each viewpoint.

\subsection{Color-INR}

In this subsection, we introduce the details of \textbf{Color} \textbf{I}mplicit \textbf{N}eural \textbf{R}epresentations (Color-INR), where the 3D structure information of the scene is encoded into the density attribute and the color information is encoded into the color attribute. 

The Color-INR represents a scene as a neural volume field by sampling points along the rays that pass through the center of the camera and each pixel in the input image (see detail in Fig. \ref{method}). The position ($x,y,z$) of each sampled point and view direction ($\theta$, $\beta$) of each ray are encoded by a positional encoding layer and then feed into a set of spatial-MLPs $\Phi_{s}$ and direction-MLPs $\Phi_{d}$:
\begin{equation}
    c,\sigma = \Phi_{d}(\Phi_{s}(x,y,z),\theta,\beta)),
\end{equation}
where the output $c$ and $\sigma$ are the color attribute and density attribute respectively. The color attribute represents the $(R, G, B)$ component and the density attribute represents the probability of light passing through this point, which can be considered as the weight of the color attribute and can be further processed to the depth attribute or the mesh results. Subsequently, we render all the density and color attributes along a ray and get the ray-color value $\hat{C}(r)$ by a discretized volume fraction:
\begin{equation}\label{eq:rendering}
\hat{C}(r) = \sum \limits ^{N}_{i=1} \exp{\Big(-\sum \limits ^{i-1}_{j=1}\alpha_j\sigma_j\Big)}\Big(1-\exp{(-\alpha_i\sigma_i})\Big)c_i,
\end{equation}
where $\alpha_i$ is the interval distance between the $i$ point and the $i+1$ point. $\hat{C}(r)$ is the predicted color of the pixel on the image space corresponding to the ray $r$. We optimize the weights of MLPs by minimizing the sum of the square difference of all input pixel values and all rendered output:
\begin{equation}
\mathcal{L}_{rgb} = \sum \limits_{r \in R}\Big[\Vert \hat{C}(r)-C(r)\Vert_2^2\Big],
\end{equation}
where $R$ is the set of all the sampling rays. After continuously optimizing the color attribute and density attribute of the sampling points in the rays, we finally get an implicit neural representation of the target scene. The geometry information of the scene is decoded into the density attribute of all sampled points. For example, we can render the depth map or the mesh result~\cite{yuan2022nerf} only using the density attribute, as shown in Fig.~\ref{Fig1}(c). Due to the continuity of the spatial attributes in the high-dimensional space, we can also render novel view images that are not included in the training views.


\subsection{Seg-INR}

In the proposed Seg-INR, we utilize the 3D structure prior, encoded in the point features, to achieve novel view segmentation. For the simplicity of the network structure design, we freeze the weights of $\Phi_{s}$ and then convert the 3D structure prior (the point features extracted in $\Phi{s}$) to the ray-semantic features. This operation also avoids repeated training of the $\Phi_{s}$ when adding new annotations. However, this will add difficulties to make full use of the texture in RGB images and complete the semantic information from unseen viewpoints. In addition, the color attributes change with the viewing angle, but the semantic attributes of points along a ray are ideally consistent from different viewing angles.

To address the above problems, we first eliminate the direction information $(\theta, \beta)$ and then design a memory-friendly Ray-Transformer to integrate the density features. Specially we take the CNN features as an additional token into the Ray-Transformer to broadcast the texture information. Finally, we combine the CNN features with ray-semantic features and feed the results to a Seg-Header to produce the final segmentation output.

\begin{figure}[t]
\centering
\includegraphics[width=\linewidth]{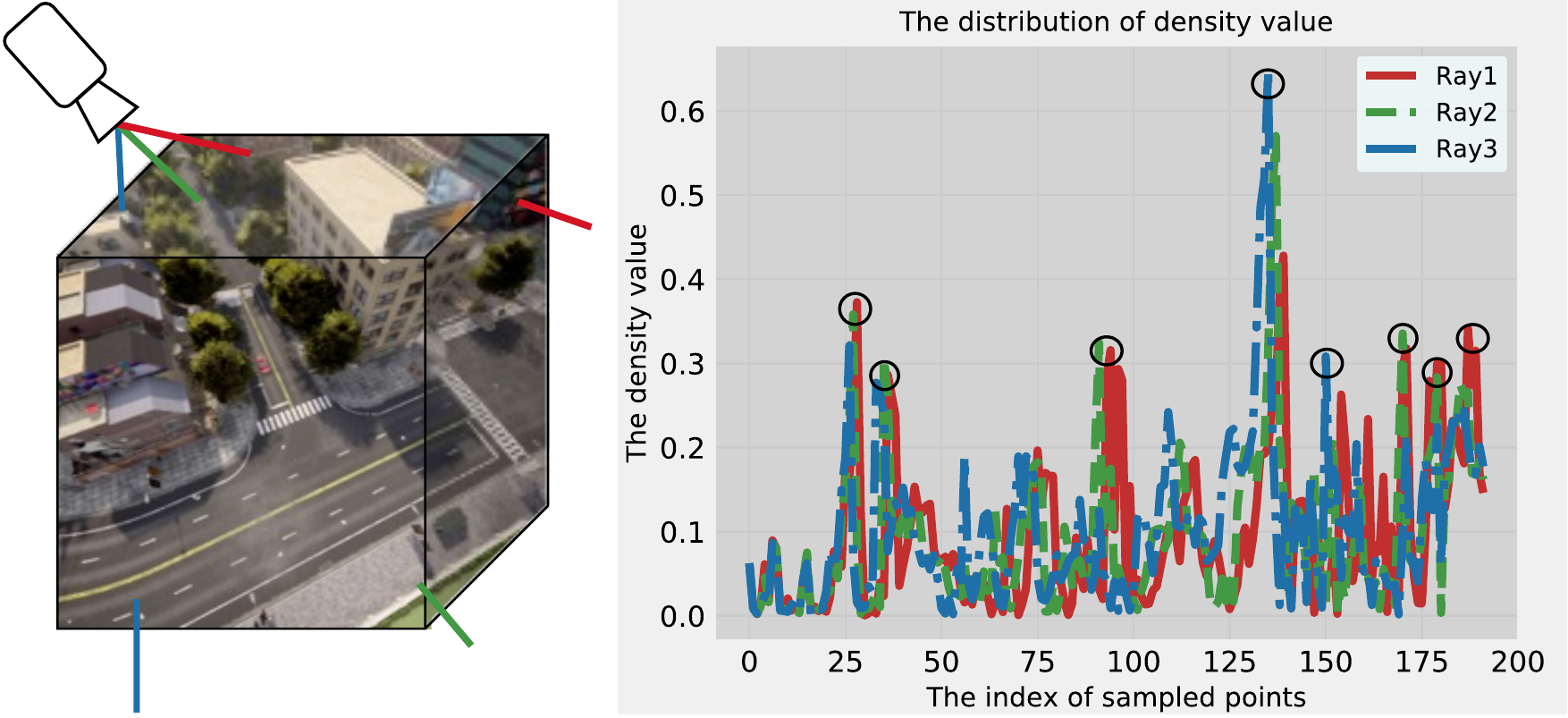}
\caption{The right image shows the three rays we randomly select, and the left image is the density curve of the three rays. We can find that the density of most points is relatively low, which represents points in the object-free space. These points contribute very little to the result. The points in the black circles represent valid points near the objects.}
\label{density}
\end{figure}

\textbf{Memory-friendly Ray-Transformer:}
The long-range attention in a transformer is efficient for integrating the features of points along a ray to keep semantic consistency. However, integrating all the point features along a ray will cause two drawbacks (Fig. \ref{density}): 1) The points with a low-density value in object-free space contribute less to the color-class conversion and has a non-uniform semantic attribute in different rays. The points in object-free space may have different semantic attributes from different view directions; 2) The computational complexity of the transformer will be greatly increased by adding more input tokens. Considering the above drawbacks, we design a memory-friendly transformer to effectively integrate and transfer the color features into semantic features.

In the memory-friendly transformer, a density-based selector is designed which firstly chooses k (k$<$ the number of sampled points) valid points along a ray and then feeds them into a transformer encoder, which consists of multi-head self-attention (MSA) layers and MLP blocks. Specially, we only integrate the 3D context relationship along a ray instead of in the $h\times w$ plane. At each layer $l$, the query, key, and value features of each selected valid point feature $F$ are extracted by the MLP blocks:
\begin{equation}
\begin{aligned}
    &Q_l = F\times W^q_l,\\
    &K_l = F\times W^k_l,\\
    &V_l = F\times W^v_l,
\end{aligned}
\end{equation}
The self-attention at layer $l$ is formulated as:
\begin{equation}
    {\rm Att}(Q_l,K_l,V_l) = {\rm softmax}\left(\frac{Q_lK_l^T}{\sqrt{d}}\right)V_l,
\end{equation}
where $d$ is the dimension number of features. 

The self-attention mechanism globally considers all the input point features, which is suitable for keeping the consistency of semantic attributes of points along a ray. The core idea of the transformer encoder is the multi-head self-attention (MSA) which explores the multiple-mode relationship to make the model more robust. The MSA performs multiple independent self-attention heads in parallel, the outputs of heads are concatenated and then projected into the final semantic feature of each point.

\begin{equation}
\begin{aligned}
    &{\rm MSA(F)} = {\rm concat}({\rm head_1,head_2,...,head_n}),\\
    &{\rm head_j} = {\rm Att}(Q_l,K_l,V_l)_j
\end{aligned}
\end{equation}

The point feature integration operation in the vanilla transformer is more effective for the seen viewpoints but not efficient enough for unseen viewpoints. In order to complete the texture information from unseen viewpoints, we further design a CNN module: $\Phi_{cnn}$ as a texture extractor. We take the CNN feature point corresponding to the training ray as an additional texture token to broadcast the texture information to other tokens in the transformer
\begin{equation}
s_1, s_2, \dots, s_n = {\rm RT}(r_1, r_2, \dots, r_n, c)
\end{equation}
where $r_1, \dots, r_n$ are the INR features and $c$ is the CNN feature. RT is the Ray-Transformer. We name this combination of networks as Ray-Transformer with Texture (RTT) (Fig.~\ref{method}). 

After that, a semantic feature-render takes the converted features as input and the density value from frozen $\Phi_{s}$ as the weights and renders the final ray-semantic features of the training ray:
\begin{equation}
\hat{{S}_s}(r) = \sum \limits ^{N}_{i=1} \exp{\Big(-\sum \limits ^{i-1}_{j=1}\alpha_j\sigma_j\Big)}\Big(1-\exp{(-\alpha_i\sigma_i})\Big)s_i,
\end{equation}
where $s_i$ is the points feature from the Ray-Transformer, $\alpha_i$ is the valid alpha after select layer. With the ${\hat{S}_s}(r)$, we can simply feed it into a seg-head to get the class prediction.


\textbf{Semantic information completion:} After the semantic information is introduced into the INR space by RTT, we explore another way to further complete the semantic information in novel viewpoints.
We combine the CNN features from $\Phi_{cnn}$ with the ray-semantic features ${\hat{S}_s}(r)$ and we select the feature points corresponding to the training rays to align with the ray-semantic features:
\begin{equation}
    C_s(r) = {\rm select_r}(\Phi_{cnn}(RGB)).
\end{equation}
The $\Phi_{cnn}$ is supervised by the labels and generalizes to unlabeled viewpoints. The $\Phi_{cnn}$ can complete the missing information in the novel viewpoints, which is important to generate detail-rich results. Finally, we feed the fused features into a seg-head that consists of a linear prediction layer:
\begin{equation}
\begin{aligned}
    &\hat{{S}_c}(r) = {\rm concat}({\hat{S}_s}(r), C_s(r)),\\
    &\hat{S}(r) = {\rm Seg_H}({\hat{S}_c}(r))
\end{aligned}
\end{equation}
With the 3D ray-semantic features, the $\Phi_{cnn}$ is easier to learn robust features using a simple structure. The experiments show that the further combination can greatly improve the performance of the model.

\subsection{Implementation Details}

\textbf{Network Details.} In the color-INR, we adopt the two-stage point sampling strategy by following NeRF++~\cite{zhang2020nerf++}. In the first coarse sampling, we take 64 training points uniformly on the training ray. In the fine stage, we sample 192 points based on coarse points by importance sampling to make the points more concentrated near the object. In seg-INR, we only convert the features in the fine stage to accelerate training and inference. In the Ray-Transformer, we set the layer number of the transformer to 2 and set the number of valid points in a ray to 10.

\textbf{Loss function.} 
In the Color-INR, we minimize the $\mathcal{L}_{rgb}$ to optimize the $\Phi_{s}$ and $\Phi_{d}$. In the Seg-INR, we minimize the cross-entropy loss $\mathcal{L}_{s}$ to optimize the Ray-Transformer parameters. Specially, we add an extra loss $\mathcal{L}_{cnn}$ to ensure the accuracy of CNN features. Formally, the loss function is defined as:
\begin{equation}
    \mathcal{L}_{seg} = \mathcal{L}_{s} + \mathcal{L}_{cnn},
\end{equation}
where the $\mathcal{L}_{s}$ and $\mathcal{L}_{cnn}$ are defined as follows:
\begin{equation}
    \mathcal{L}_{s} = -\sum \limits_{r \in R} \Big[ \sum \limits_{l=1}^{L} S^l(r)\log{\hat{S}}^l(r)\Big],
\end{equation}
\begin{equation}
    \mathcal{L}_{cnn} = -\sum \limits_{r \in R}\Big[\sum \limits_{l=1}^{L} S^l(r)\log{C_s}^l(r)\Big],
\end{equation}
where $L$ is the number of classes.

\section{Experimental Results and Analysis}\label{section:experiments}

\subsection{Dataset and Metrics}

In our experiment, we construct a large and challenging dataset for the multi-view remote sensing image segmentation task. In our dataset, we collect six synthesis sub-datasets using the CARLA platform and three real sub-datasets from Google Maps. The dataset contains 865 images and the size of each image is 512 $\times$ 512 pixels. Only 2\% to 6\% of the images in the training set have corresponding segmentation labels. The dataset contains scenes at multiple scales, varying from a number of buildings to an entire town. Table~\ref{datasetdescription} and Table~\ref{classdescription} show the dataset statistics and their category distribution. Among them, due to the automatic labeling of the CARLA platform, the labels in the synthetic data set are more refined. 

\begin{table*}[!htb] \small
\centering
\caption{Category distribution of the synthetic and real dataset}
\label{classdescription}.
\begin{tabular}{c|ccccc}
\toprule
Dataset              & \multicolumn{5}{c}{Class}                                           \\ \midrule
\multirow{4}{*}{SYSs} & Building:43.46\%  & Fence:0.11\%  & Other:0.73\%  & Pole:0.45\%  & RoadLine:0.72\%  \\ 
                     & Road:12.93\%  & Sidewalk:17.91\%  & Vegetation:8.32\%  & Vehicles:0.20\%  & Wall:0.10\% \\
                     & Traffic-Sign:0.01\% & Sky:0.55\% & Ground:1.75\% & Bridge:0.17\% & RailTrack:1.47\% \\
                     & Traffic-Light:0.01\% & Static:1.03\% & Dynamic:0.14\% & Water:6.70\% & Terrain:1.80\% \\ \midrule
REAL1                 & Background:57.80\%  & Building:40.86\%  & Car:0.07\%  & Grass:0.02\%  & Tree:0.25\% \\\midrule
REAL2                 &   & Background:46.92\%  & Bridge:25.85\%  & Building:26.22\%  &  \\\midrule
REAL3                 &   & Background:49.68\%  & Building:35.48\%  & Road:13.84\%  &  \\\bottomrule
\end{tabular}
\end{table*}

\begin{table}[!htb] \small
\centering
\caption{Details of the synthetic and real dataset.}
\label{datasetdescription}.
\begin{tabular}{c|cccc}
\toprule
         & \#views & \#annoations & \#classes & \#size (pixels) \\ 
\midrule
sys \#1 & 100   & 3/3.0\%  & 20     & $512\times 512$ \\ 
sys \#2 & 100   & 4/4.0\%  & 18   & $512\times 512$ \\ 
sys \#3 & 100   & 5/5.0\%  & 20   & $512\times 512$ \\ 
sys \#4 & 80    & 5/6.3\%   & 19 & $512\times 512$ \\  
sys \#5 & 85    & 5/6.0\%   & 19 & $512\times 512$ \\  
sys \#6 & 100   & 5/5.0\%   & 18 & $512\times 512$ \\  
real \#1 & 100  & 2/2.0\%   & 5 & $512\times 512$ \\  
real \#2 & 100  & 2/2.0\%   & 3 & $512\times 512$ \\  
real \#3 & 100  & 2/2.0\%   & 3 & $512\times 512$ \\  
\bottomrule
\end{tabular}
\end{table}

\begin{figure}[!htb]
\centering
\includegraphics[width=\linewidth]{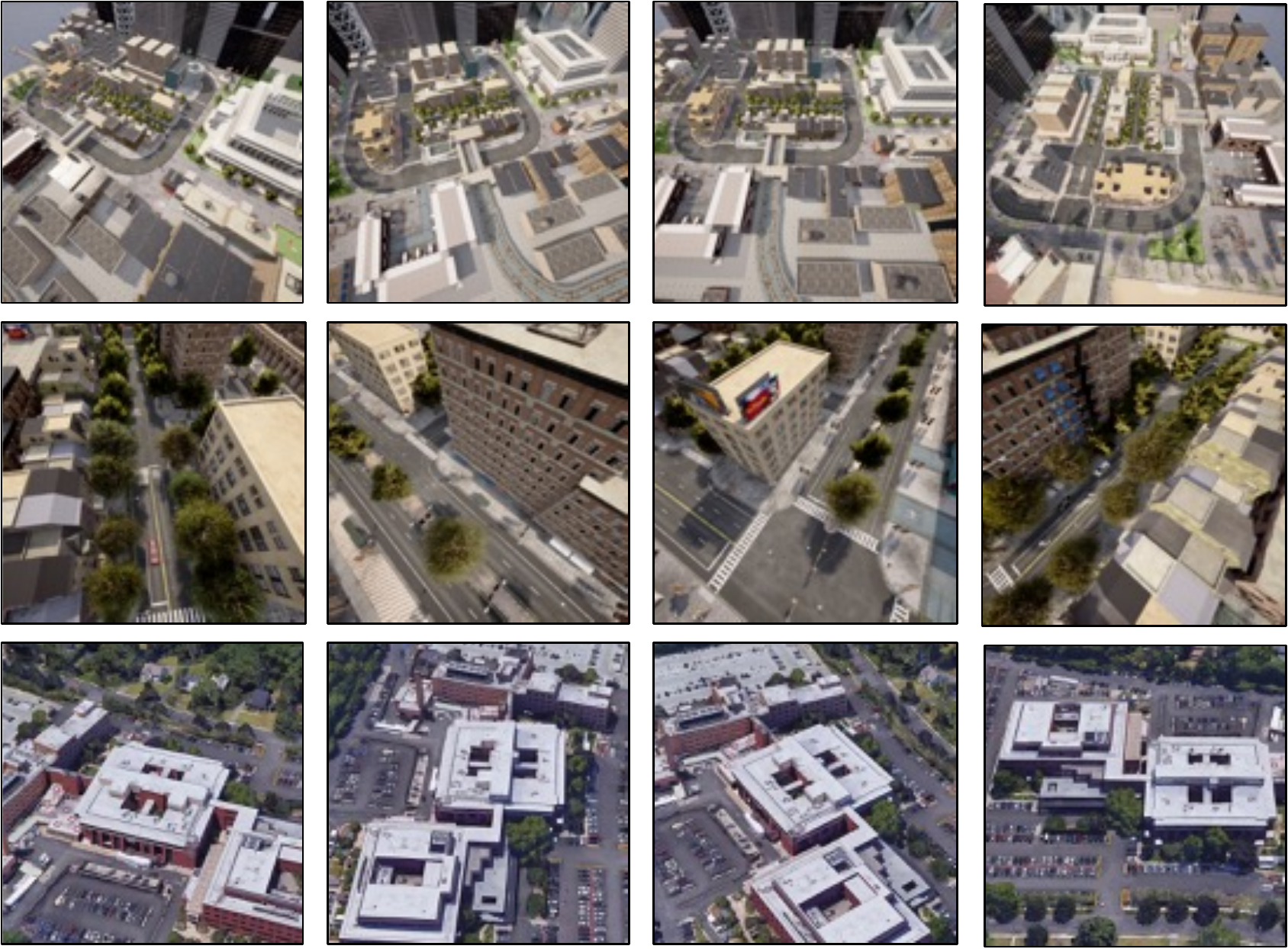}
\caption{The first row is the data samples in sys \#1; The second row is the data samples in sys \#2; The last row is the data samples in real \#1.}
\label{datasample}
\end{figure}

Our dataset poses three challenges to the multi-view image segmentation task. The first challenge is the limited annotation, where the labeled images only occupy 2\% - 6\% of the whole training set. The second challenge is the category imbalance problem, as shown in Table~\ref{classdescription}. The third challenge is that different ground objects may have similar textures and contain some densely located ground objects, which are difficult to distinguish, as shown in Fig.~\ref{datasample}. We can find that building roofs and roads have similar textures, and there are also some densely parked cars.


In our experiment, we use mean Intersection over Union (mIoU), the most commonly used segmentation metric, to compare the performance of different methods:
\begin{equation}
    mIoU = \frac{TP}{FP+FN+TP}
\end{equation}
where the $TP$ is the true positive, $FP$ is the false positive, $FN$ is the false negative.

\subsection{Comparison Methods}

In order to verify the effectiveness of our method, we choose the following methods for comparative experiments. All comparative methods are retrained on the proposed dataset to ensure fairness. Besides, we also compared different strategies to introduce texture information into the INR space. The baseline strategy is to take color reconstruction and semantic segmentation as a multi-task \cite{zhi2021place}. Another two-stage strategy is to reconstruct color information first and then convert it into segmentation \cite{qi2022remote}. We also compare three different approaches to using a transformer to integrate point features based on the above two-stage strategy. The first approach is to use a transformer to simply integrate long-range point features along a training ray to obtain semantic features; the second approach is to design a spatial CNN token and broadcast texture information from an input RGB image into other tokens; the final way is to further concatenate the CNN features with integrated features by the transformer to enhance the detail texture information. In addition to the above variants, we also conducted a horizontal comparison with methods published in recent years. The comparison methods are as listed as follows:



\begin{enumerate}
\item SegNet~\cite{badrinarayanan2017segnet}: a CNN-based semantic segmentation method that uses encoder-decoder architecture;
\item Unet~\cite{ronneberger2015u}: a CNN-based semantic segmentation method that adopts the skip-connection to keep the detail information in encoder layers;
\item DANet~\cite{fu2019dual}: a CNN-based semantic segmentation method that proposes a dual attention mechanism to adaptively integrate local features and global dependencies;
\item DeepLabv3~\cite{chen2017rethinking}: a CNN-based semantic segmentation method that uses the dilated convolution to effectively enlarge the receptive field; 
\item SETR~\cite{zheng2021rethinking}: a transformer-based semantic segmentation method;
\item Sem-NeRF~\cite{zhi2021place}: the first NeRF-based model for indoor image semantic segmentation;
\item Color-NeRF~\cite{qi2022remote}: a NeRF-based method for semantic segmentation that adds an additional color-radiance network to fuse the pixel-level color information for improving the NeRF-based segmentation;
\item IRT (B): Our baseline method which transforms the color features into semantic features only by fine-tuning;
\item IRT ($RT_2$): A variant of our method that integrates the point features along a ray based on the baseline by a two-layer transformer;
\item IRT ($RT_6$): A variant of our method that has a similar configuration with IRT ($RT_2$) except the number of transformer layers is set to 6;
\item IRT (RTT): A variant of our method where we take the CNN features as semantic guidance based on IRT ($RT_2$);
\item IRT (RTC): A variant of our method where we combine the ray features and CNN features produced by the Ray-Transformer based on IRT ($RT_2$);
\item IRT (RTTC): A variant of our method where we combine the ray features and CNN features produced by the RTT.
\end{enumerate}

\subsection{Overall Comparison Results}

\begin{figure*}[!htb]
\centering
\includegraphics[width=\linewidth]{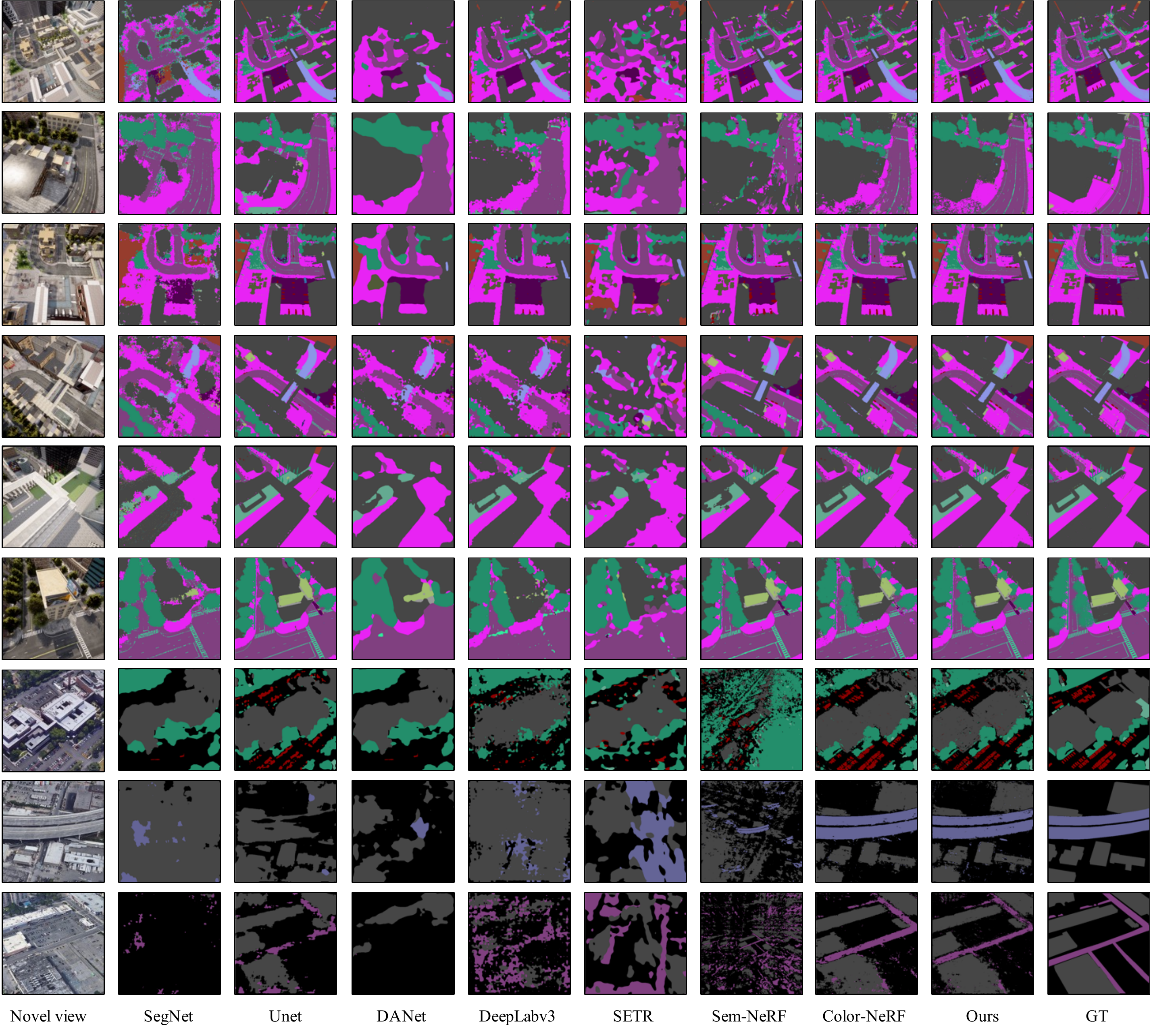}
\caption{In the CNN-based methods, Unet has the best performance but it still fails in real scenes. Compared to other methods, our method generates more accurate and complete results, such as zebra crossing, small trees, and traffic lights. The advantage is the introduction of texture information from the Ray-Transformer. At the same time, our method achieves more robustness in the multi-scale object. This is due to the continuity of spatial attributes in the high dimension. From top to bottom, the images of each row are from scenes sys \#1, sys \#2, sys \#3, sys \#4, sys \#5, sys \#6,
real \#1, real \#2, real \#3}
\label{normal-lighting-results}
\end{figure*}

\textbf{Qualitative and Quantitative Results}
In Fig.~\ref{normal-lighting-results}, we show the qualitative results of CNN-based and INR-based methods. The results show that the INR-based methods get better results than the CNN-based methods. Although Unet has the best performance in all CNN-based methods, it still fails in the dataset real \#2. The key reason is that CNN-based methods that only emphasize 2D information, which relies more on a large number of annotations. It is not enough to distinguish similar objects with the features that are extracted from a limited number of samples. For example, the building class is more likely to be predicted to the road in the dataset sys \#2, and the bridge or road is easily misclassified in real scenes. On the contrary, our method with the implicit 3D information from Color-INR is more powerful to distinguish similar textures for various scenes. The key advantage is that we take the 3D structure into consideration to help the model distinguish the objects in different depths. Compared to Sem-NeRF, our method can generate more detailed and clearer segmentation of small-scale objects, such as zebra crossing, small trees, and traffic-light. Our method also successfully achieves dense object segmentation in real \#1. At the same time, our method can generate complete and accurate big-scale object segmentation, such as the building in syn \#5. The adaptability of our method at multiple scales is due to the continuity of high-dimensional spatial properties, which do not depend on the resolution. It is noticed that Sem-INR fails in real sub-datasets. This is due to the poor ability of the vanilla NeRF-based method to reconstruct the scenes with larger differences between foreground and background. 

From Table~\ref{metric-results}, we can find that the INR-based methods get a higher mIoU score than CNN-based methods in all sub-datasets. Compared to the Unet which has the best performance in CNN-based methods, the proposed IRT outperforms the Unet by 15.5\% on the average mIoU. Compared to the Sem-INR, our method outperforms it by 1.93\%, 8.42\%, 3.65\%, 2.74\%, 2.29\%, 6.74\%, 54.2\%, 65.53\%, 51.26\% respectively, in each sub-dataset. Our results also outperform the Color-NeRF by 1.81\% in average mIoU. Our method shows greater superiority on synthetic data, suggesting that IRT is more robust to the cases with more classes and detailed annotations. Further comparisons along synthetic sub-datasets show that with the introduction of texture information, our model further improves on generating detailed results for high-resolution images, \textit{e.g.} sys \#2. 

\begin{table*}[!htb] 
\centering
\caption{Quantitative comparison of different methods.}
\label{metric-results}
\begin{tabular}{c|cccccccccc}
\toprule
Methods  & sys \#1 & sys \#2 & sys \#3 & sys \#4 & sys \#5 & sys \#6 & real \#1 & real \#2 & real \#3 & AVG \\\midrule
SegNet &11.79& 13.21 & 10.81 & 26.71 & 8.21 & 18.77 & 8.67 & 27.88 & 18.84 & 16.10 \\ 
Unet &23.92  & 31.73 &42.26 & 41.79 & 26.63 & 38.72& 64.94 & 49.95& 68.33& 43.14   \\ 
DANet  & 9.73 & 13.91 & 16.62 &  26.78 & 8.53 & 15.79 & 35.71& 49.13& 34.30&  23.39   \\ 
Deeplab  & 18.89 & 16.64 & 19.96 & 30.42 & 11.90& 20.52 & 39.35 & 41.37& 52.37 & 27.94    \\ 
SETR & 10.34&10.71 &12.45 &21.13& 8.90 & 13.97& 36.26 & 31.04 & 26.07 & 18.99  \\ 
Sem-NeRF  & 55.73 & 34.81& 49.82 &57.24& 41.44 & 41.85 & 11.64 & 18.78& 19.76 &36.79 \\ 
Color-NeRF & \textbf{58.03} &38.46 & 50.86 & 59.14 & \textbf{43.77} & 43.53 & 62.04 & 85.85 & 69.92 & 56.84\\ \midrule
IRT(B)  & 53.12 & 33.95 & 44.69 & 55.30 & 39.08 & 41.20 & 60.65& 83.40 & 69.37 & 53.42 \\ 
IRT($RT_2$) & 53.70 & 34.97 & 44.97 & 55.47 & 39.09 & 41.50 & 60.96 &83.58 & 70.98 &53.91 \\
IRT($RT_6$) & 52.70 & 34.97 & 44.60 & 55.65 & 39.57 & 42.31 & 60.51&  83.57  & 71.37  &53.92 \\
IRT(RTT) & 54.39&41.65&51.79&57.98& 41.28 &46.99& 65.61 &   85.98&\textbf{71.71} &57.49\\
IRT(RTC)   & 57.61 & 42.19& 49.43 & \textbf{60.33} & 42.43 & 45.15 & 65.51 & \textbf{86.06} &62.06 & 56.75\\
IRT(RTTC)&57.86 &\textbf{43.23}&\textbf{53.47}  &59.98& 43.73 & \textbf{48.59} & \textbf{65.84} & 84.31& 71.02&\textbf{58.67}\\ \bottomrule
\end{tabular}
\end{table*}

\begin{figure}[!htb]
\centering
\includegraphics[width=1.0\linewidth]{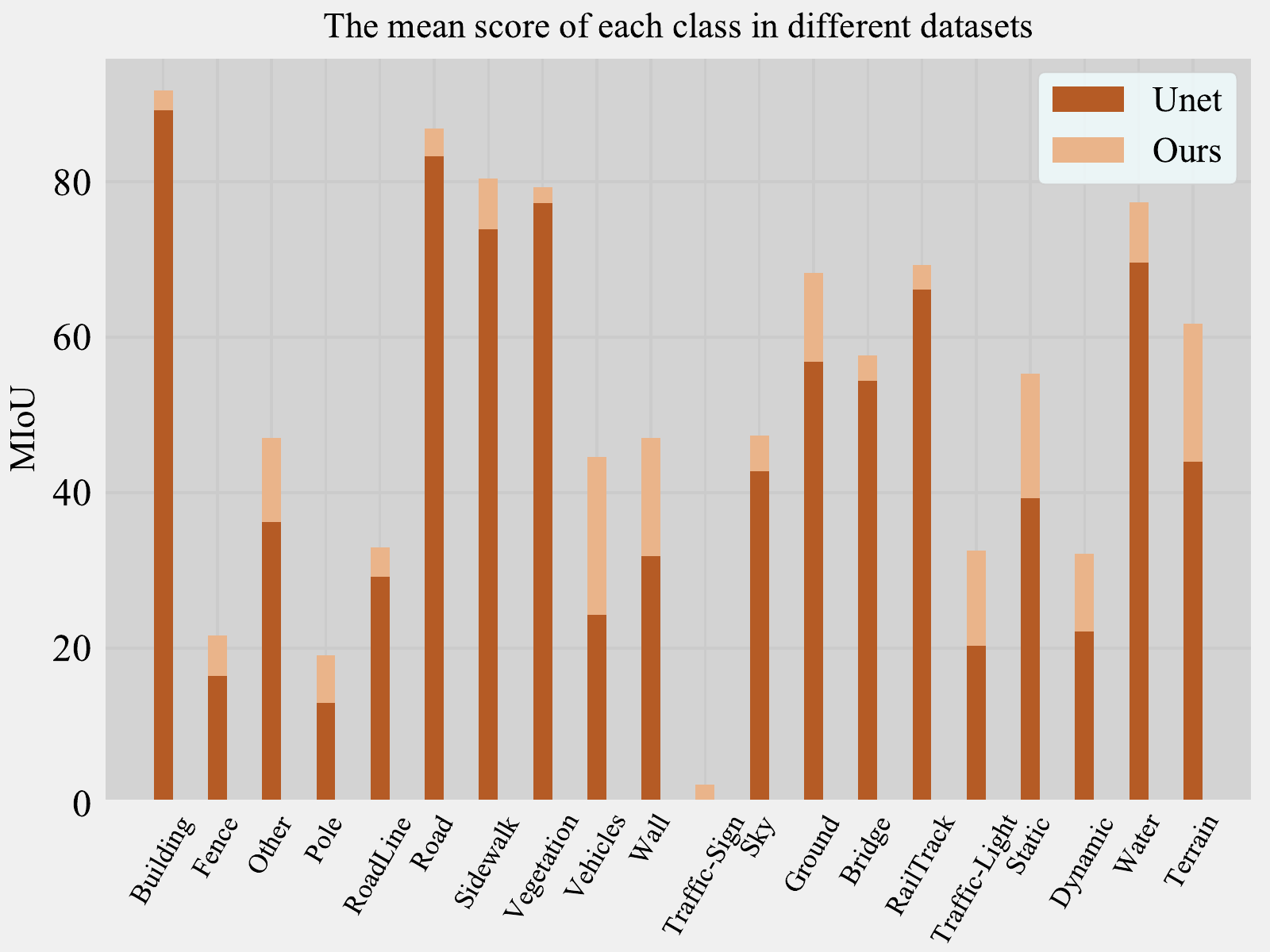}
\caption{The avg mIoU of each class in sub-datasets. The result shows that the proposed method is more friendly to the classes with a small number of annotations.}
\label{eachclass}
\end{figure}

\begin{figure}[!htb]
\centering
\includegraphics[width=\linewidth]{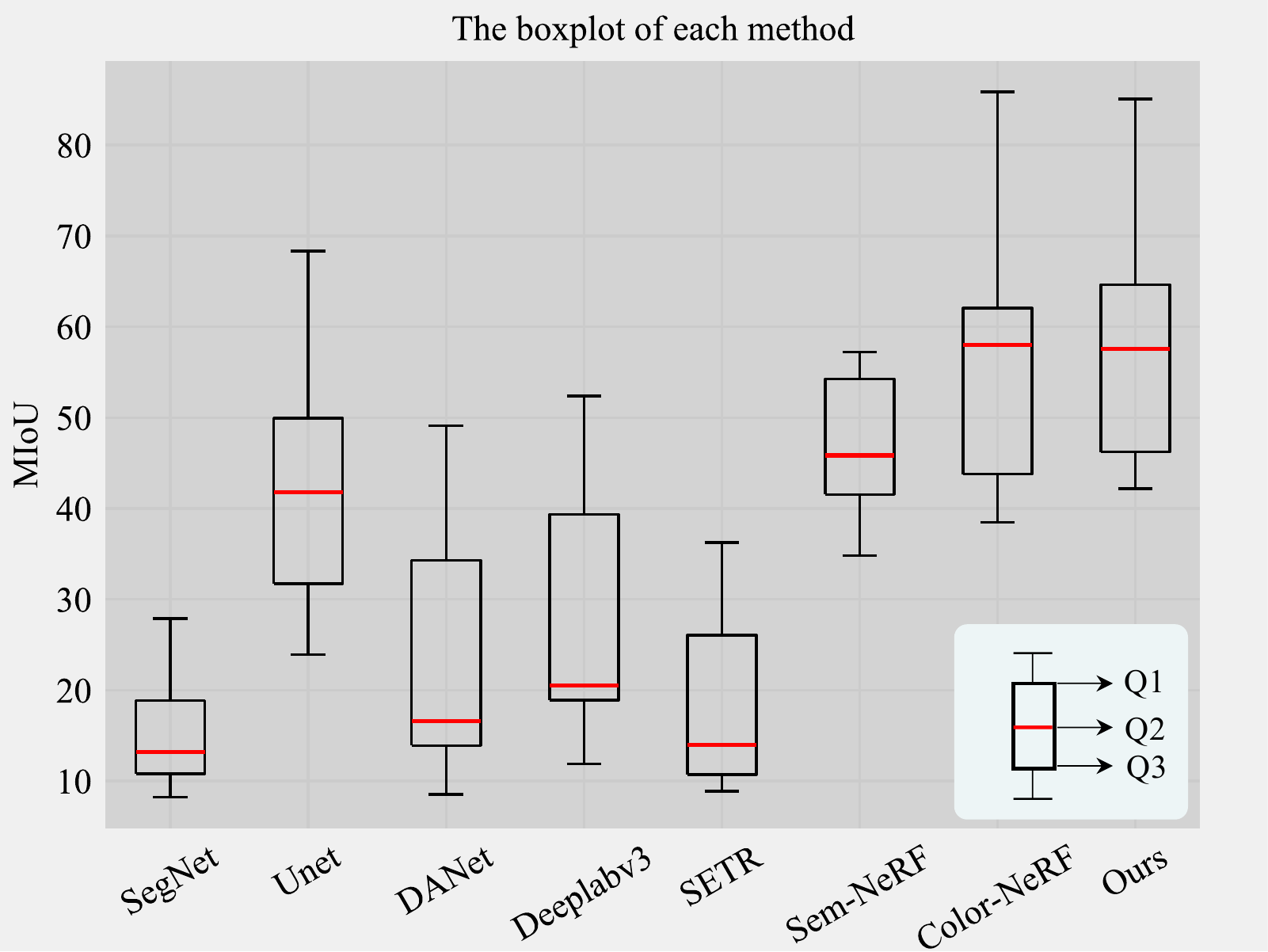}
\caption{The boxplot results of all methods. The Q1, Q2, and Q3 values of our method are higher. Q1 : first quartile value of the result; Q2: the median
value of the result; Q3: the last quartile value of the results.}
\label{compairedbox}
\end{figure}

In Fig.~\ref{eachclass} and Fig.~\ref{compairedbox}, we show the statistical results at the category level and dataset level. Fig.~\ref{eachclass} shows the average mIoU across all synthetic sub-datasets for each category. The results show that the category with a large number of samples usually has a higher value of mIoU. We can also find that the IRT has more advantages in classifying categories with a small number of samples and even with only a few pixel labels (\textit{e.g.}, Traffic-Sign category). Fig.~\ref{compairedbox} shows the boxplot results of mIoU distribution for each sub-dataset. The results show that the Q1, Q2, and Q3 values are obviously higher than others, and our method has greater potential for better results.(Q1 : first quartile value of the result; Q2: the median value of the result; Q3: the last quartile value of the results. The boxplot reflects the scatter of results )

\begin{figure*}[!htb]
\centering
\includegraphics[width=\linewidth]{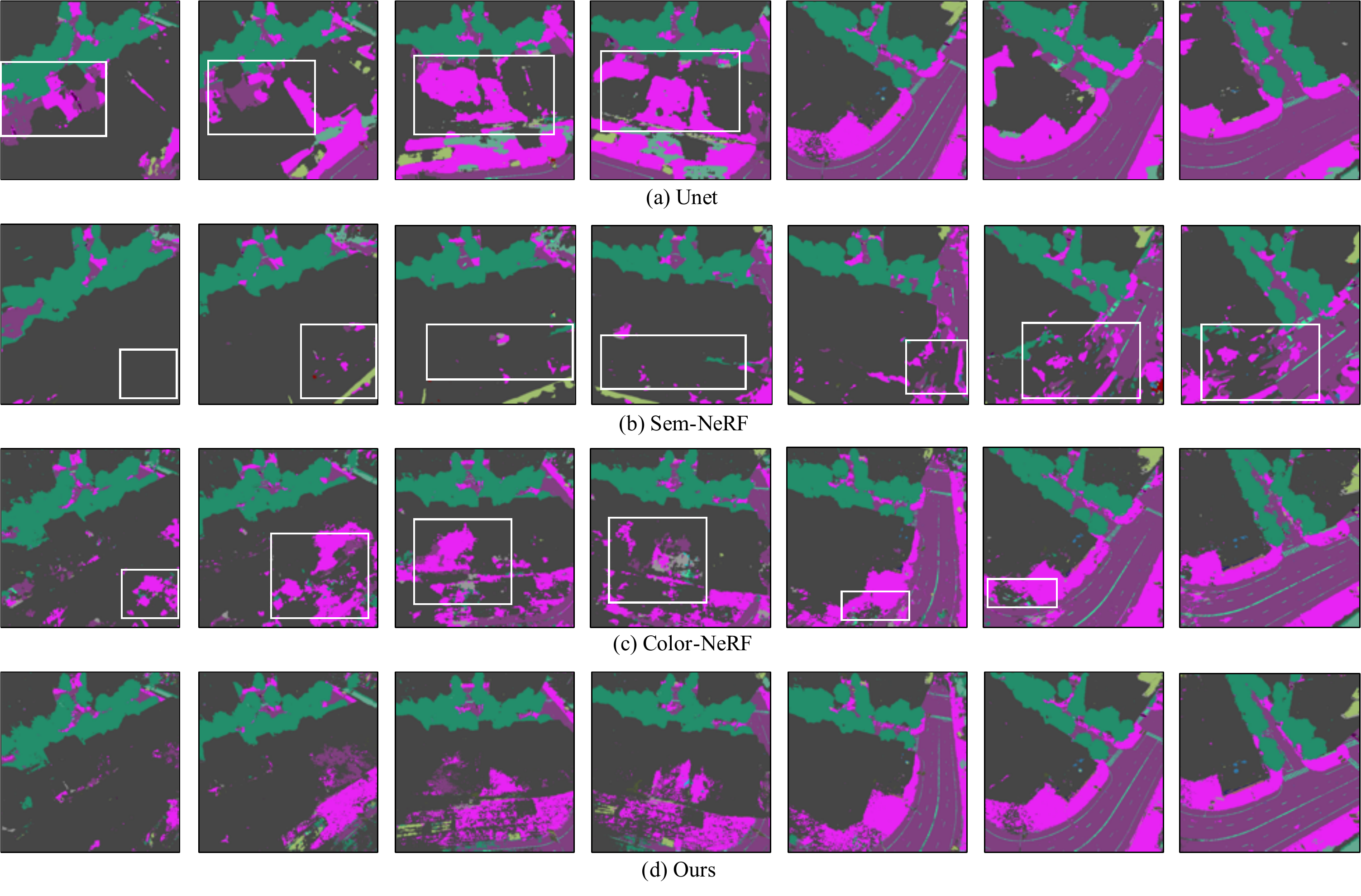}
\caption{The semantic-consistent results from sys \#2. The results show that IRT can generate results with more accuracy and view consistency. The labels in white boxes are not accurate or view-consistency.}
\label{view-consistency}
\end{figure*}

\textbf{Semantic Consistency}. We compared the consistency of the segmentation results of different methods, including Unet, Sem-INR, Color-INR, and ours. In Fig.~\ref{view-consistency}, we can see that Unet has low accuracy in classifying buildings from the first few viewpoints, and since it processes images from each viewpoint separately, it does not maintain view consistency.  On the contrary, our methods work in the 3D INR space and mainly take the location of spatial points as the input, therefore, presenting better view consistency.   Besides, since our method incorporates CNN features, it can achieve more accurate results compared to Color-NeRF which only extracts pixel color information.

\subsection{Ablation Study}

\begin{figure}[!htb]
\centering
\includegraphics[width=\linewidth]{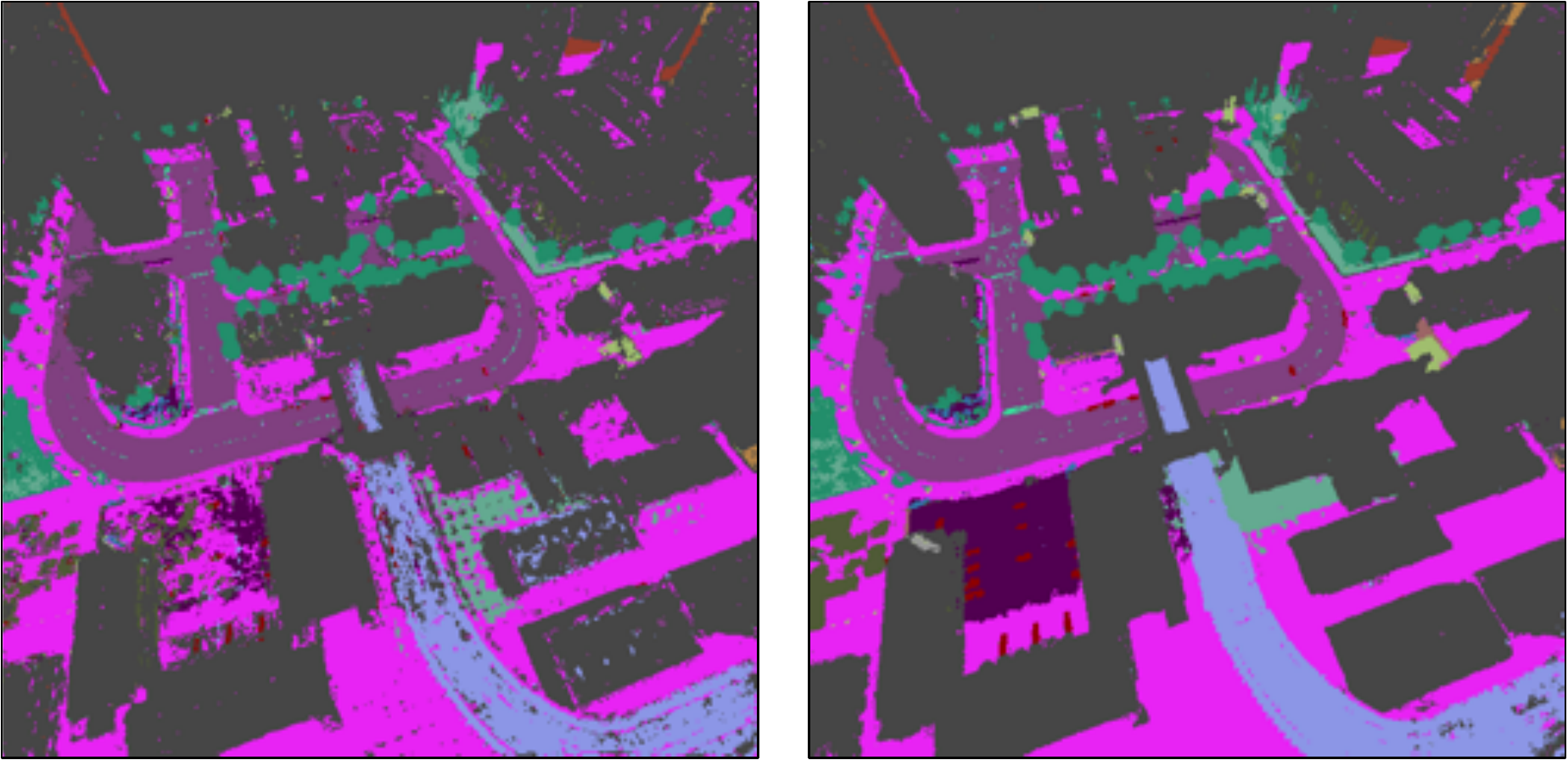}
\caption{The left image shows the result without Color-INR and the right image shows the result with Color-INR. Compared the two results, we can find that the Color-INR which encodes the 3D structure of the scene can effectively keep the spatial-semantic continuity of the result. }
\label{without}
\end{figure}

In this part, we conduct ablation experiments, mainly
to verify: 1) the necessity of the CNN token, 2) the effectiveness of the number of valid points in the Ray-Transformer, and 3) the necessity of the Color-INR. Our evaluation results are shown in Tab.~\ref{points} and the lower part of Tab.~\ref{metric-results}. The visual results are shown in Fig.~\ref{without}. 

\textbf{Ray-Transformer and CNN token.} In Tab.~\ref{metric-results}, the IRT ($RT_2$) and IRT ($RT_6$) represent the vallina Ray-Transformer with  different encoder layers. Compared to the baseline model, we can see that the vanilla Ray-Transformer only brings minor improvement and the deeper transformer structure can not further improve the performance. The reason is that the vanilla Ray-Transformer, which mainly integrates the long-range relationship of input token in ray space generates more view consistency and detailed results but can not further extract texture information to improve performance greatly. 

Although the vanilla transformer can not introduce extra information, its structure is suitable for fusing different information in different feature spaces, such as INR feature space and CNN feature space. The accuracy boost of IRT (RTT) suggests that Ray-Transformer taking CNN token as input is important for completing the missing information of novel view under the sparse annotations. In addition, we also try to introduce the texture features by combining the CNN feature with the vanilla Ray-Transformer features(RTC). The result is still improved, but not as much as the IRT (RTT). This demonstrates the effectiveness of the transformer architecture. 



With the combination of CNN and Transformer, the performance has been further improved and reached the SOTA. In detail, we find that adding texture features is more effective for the small-scale scenes, \textit{e.g.} sys \#2 and sys \#3. We also compare the difference between only adding the CNN token into Ray-Transformer (IRT(RTT)) with combining the CNN feature with ray-semantic features (IRT(RTC)). The results show that broadcasting texture information in ray space is more effective for small-scale scenes. The CNN features with a larger receptive field are more effective in large-scale scenes, \textit{e.g.}, sys \#1.

\textbf{The number of sampling points.} We explore the influence of the number of sampling points in the Ray-Transformer. As shown in Table~\ref{points}, we test on two different settings: 10 and 20 sampling points along a ray. We found that when we increase the point number from 10 to 20, the result improved very slightly. This implies that the points in object-free space contribute less to the final results.

\begin{table*}[!htb] \small
\centering
\caption{Quantitative comparison of using different valid points in Ray-TransFormer.}
\label{points}
\begin{tabular}{c|ccccccccc}
\toprule
 Point Number & sys \#1 & sys \#2 & sys \#3 & sys \#4 & sys \#5 & sys \#6 & real \#1 &real \#2 &real \#3 \\ \midrule
10 points & 57.86 & 43.23  & 53.47 &  59.98 & 43.73& 48.59 &65.85 & 84.31&71.02   \\ 
20 points & 57.97 &  43.28 & 53.22 &  60.18 & 43.99 & 48.67&65.00 & 84.46&71.81 \\ \toprule
\end{tabular}
\end{table*}

\textbf{The necessity of Color-INR.} 
Fig.~\ref{without} shows the results with and without the Color-INR. We retrain the model only using a few labels and discard the training results from the first stage. We can see that without the Color-INR, it is difficult for the model to obtain spatial-semantic continuity results as there are many holes in the output label map. This is because it is difficult to build volumetric scene space without the Color-INR, which means 3D information is crucial to the task.

\subsection{Robust again Illumination and View Changes}

In this part, we compare our method against CNN-based methods on the robustness against illumination and view changes.

\begin{figure}[!htb]
\centering
\includegraphics[width=\linewidth]{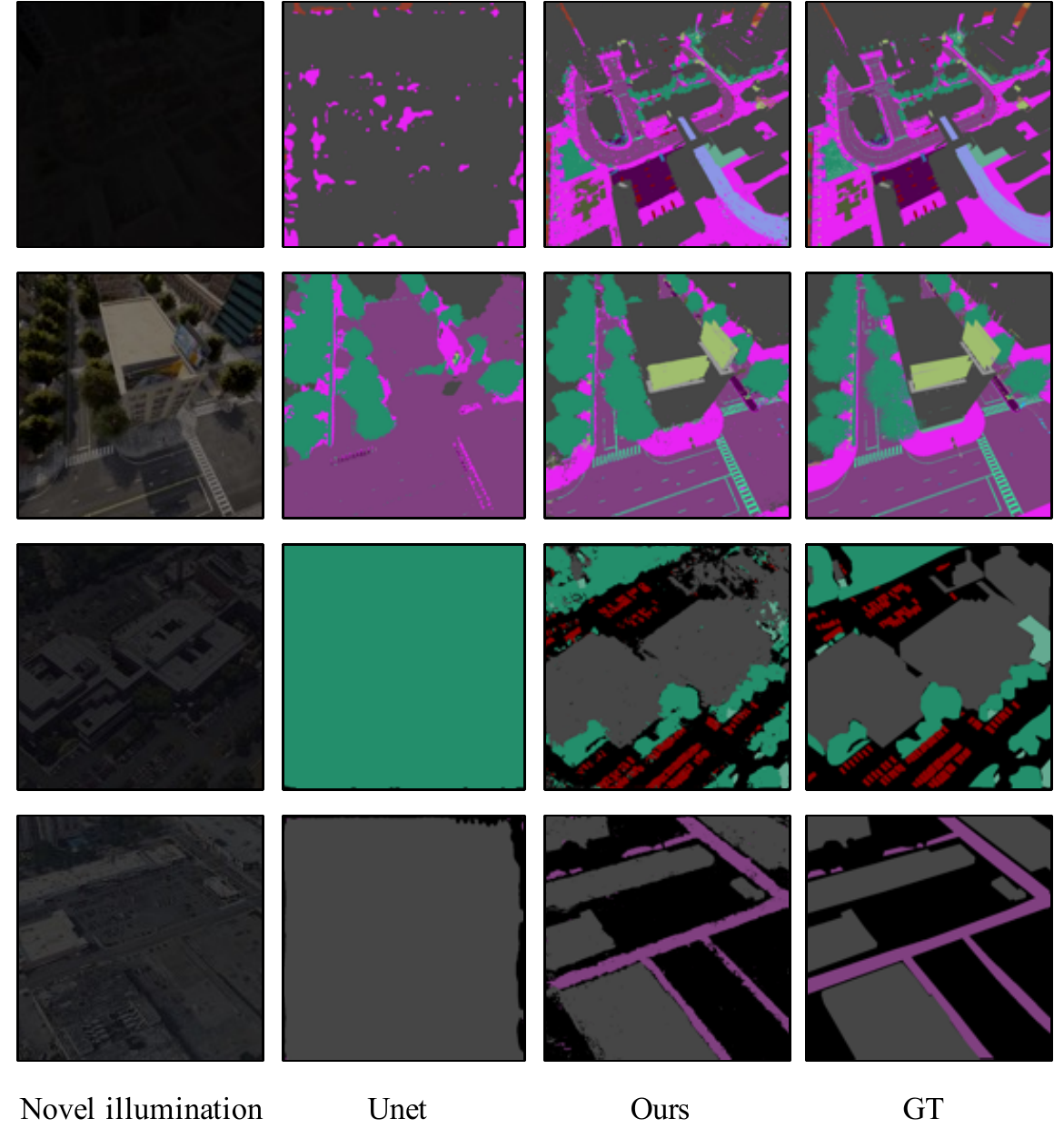}
\caption{We randomly change the intensity of the input images to simulate lighting conditions in a dark environment and test them with non-retrained models. The results show that our method significantly outperforms the CNN-based method.}
\label{dark}
\end{figure}

\textbf{Robustness against illumination changes.} We randomly change the intensity of the input images to simulate lighting changes and test directly on the non-retrained CNN-based models and our model. As shown in Table~\ref{darkresults}, the CNN-based methods show weak adaptability to illumination changes. In contrast, our method maintains the highest accuracy and the smallest precision fluctuation among all methods. The visualized result (Fig.~\ref{dark}) also suggests the above conclusion. 
The robustness of our method mainly comes from the introduced 3D structural information, which does not change with illumination.

\begin{table*}[!htb] 
\centering
\caption{Quantitative comparison under dark environment. mIoU/Performance drop(\%)}
\label{darkresults}
\begin{tabular}{c|cc|cc|cc|cc|cc|cc}
\toprule
Sub-Scenes  & \multicolumn{2}{c}{SegNet} & \multicolumn{2}{c}{Unet} & \multicolumn{2}{c}{DANet} & \multicolumn{2}{c}{Deeplab} & \multicolumn{2}{c}{SETR} & \multicolumn{2}{c}{Ours} \\\midrule
sys \#1 &  1.86 & $\downarrow$-84.22\% & 4.46& $\downarrow$-81.35\% & 3.88&$\downarrow$-60.12\% & 7.41&$\downarrow$-60.70\% & 3.12&$\downarrow$-69.82\% & \textbf{46.17}&$\downarrow$\textbf{-20.20}\% \\
sys \#2 &  3.51&$\downarrow$-73.42\% & 6.82&$\downarrow$-78.50\% & 5.32&$\downarrow$-61.75\%& 5.17&$\downarrow$-68.93\%& 4.57&$\downarrow$-57.32\% & \textbf{34.34}&$\downarrow$\textbf{-20.56}\% \\ 
sys \#3 &  2.11&$\downarrow$-80.48\% & 4.82&$\downarrow$-88.59\% & 5.21&$\downarrow$-68.65\% & 6.37&$\downarrow$-68.08\%& 4.57&$\downarrow$-63.29\% & \textbf{39.23}&$\downarrow$\textbf{-26.63}\% \\
sys \#4 &  2.42&$\downarrow$-90.93\% & 7.12&$\downarrow$-82.96\% & 8.94&$\downarrow$-66.61\% & 10.23&$\downarrow$-66.37\%& 4.58&$\downarrow$-78.32\% & \textbf{49.53}&$\downarrow$\textbf{-17.42}\% \\
sys \#5 &  3.46&$\downarrow$-57.85\% & 4.54&$\downarrow$-82.95\% & 3.17&$\downarrow$-62.83\% & 4.87&$\downarrow$-59.07\%& 3.78&$\downarrow$-57.52\% & \textbf{29.70}&$\downarrow$\textbf{-32.08}\% \\
sys \#6 &  5.40&$\downarrow$-71.23\% & 4.86&$\downarrow$-87.44\% & 5.87&$\downarrow$-62.82\% & 7.35&$\downarrow$-64.18\%& 6.63&$\downarrow$-52.54\% & \textbf{35.07}&$\downarrow$\textbf{-27.82}\% \\
real \#1 & 5.01&$\downarrow$-42.21\% & 8.29&$\downarrow$-87.23\% &15.56&$\downarrow$-56.42\% & 15.76&$\downarrow$-59.94\%&7.58&$\downarrow$-79.09\% & \textbf{58.56}&$\downarrow$\textbf{-11.05}\% \\
real \#2 & 15.20&$\downarrow$-45.48\%& 19.88&$\downarrow$-60.20\%&20.75&$\downarrow$-57.76\% & 28.10&$\downarrow$-32.07\%&19.07&$\downarrow$-38.56\%& \textbf{82.94}&$\downarrow$\textbf{-1.62}\% \\
real \#3 & 16.75&$\downarrow$-11.09\%& 15.83&$\downarrow$-76.83\%&19.18&$\downarrow$-44.08\% & 29.60&$\downarrow$-43.47\%& 9.08&$\downarrow$-65.17\%&\textbf{67.90}&$\downarrow$\textbf{-4.39}\% \\ \midrule
AVG      & 6.19&$\downarrow$-61.88\% & 8.51&$\downarrow$-80.67\% & 9.76&$\downarrow$-60.12\% & 12.76&$\downarrow$58.09\%& 7.01&$\downarrow$-62.40\%& \textbf{49.27}&$\downarrow$\textbf{-17.97}\%\\\bottomrule
\end{tabular}
\end{table*}


\begin{figure}[t]
\centering
\includegraphics[width=\linewidth]{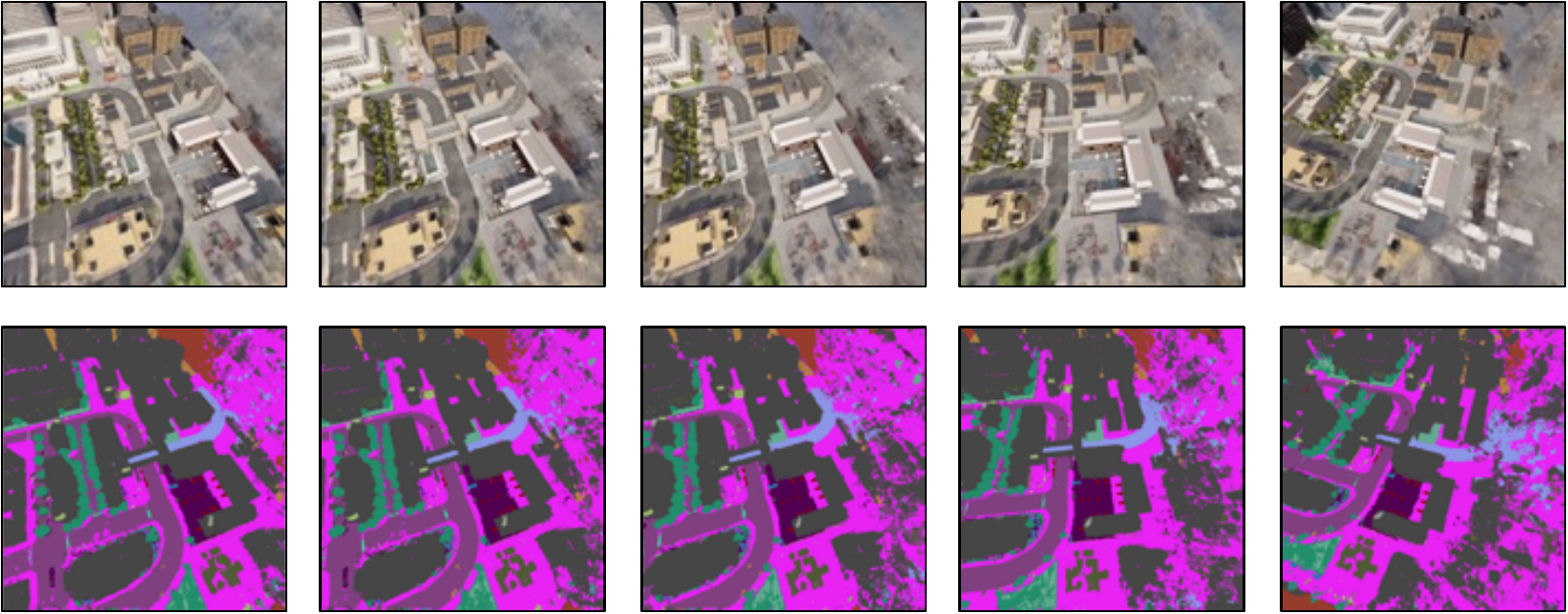}
\caption{The first row shows the images from the viewpoints that do not appear in the training viewpoints coverage. The second row shows the corresponding segmentation results.}
\label{unseen}
\end{figure}

\textbf{Robustness against view changes.} We test the robustness of our method on the novel views that were completely uncovered by the training dataset. We use the viewpoints of sys \#2 to test the model for sys \#1. Given an RGB image of the unseen view, we use the non-retrained model to produce the segmentation results. As shown in Fig.~\ref{unseen}, although our model is trained on high-altitude scenes, it can still adapt to low-altitude viewpoints and generate accurate results. At the same time, it also means that our method has greater potential in handling images of different resolutions.


\section{Conclusion}\label{section:conclusion}

In this paper, we consider multi-view remote sensing image segmentation under sparse annotations and propose a new method based on implicit neural representation and transformer. We optimize the implicit volume representation of the 3D scene by fitting the posed RGB images into a neural network. Then a Ray-Transformer network combines the CNN features with the 3D volume representation to complete the missing information of the unknown views. To achieve this, we also introduce a challenging dataset for the R4S task. Extensive experimental results verify the effectiveness of our proposed method. The results demonstrate that our method outperforms other CNN-based methods in terms of both accuracy and robustness. We also compare different strategies to add texture information into INR feature space and show the effectiveness of the transformer structure for this task. Finally, our empirical results also indicate the robustness of our method against illumination and viewpoint changes in the scene.

\ifCLASSOPTIONcaptionsoff
  \newpage
\fi

\bibliographystyle{IEEEtran}
\bibliography{refbib}

\begin{thebibliography}{10}
\providecommand{\url}[1]{#1}
\csname url@samestyle\endcsname
\providecommand{\newblock}{\relax}
\providecommand{\bibinfo}[2]{#2}
\providecommand{\BIBentrySTDinterwordspacing}{\spaceskip=0pt\relax}
\providecommand{\BIBentryALTinterwordstretchfactor}{4}
\providecommand{\BIBentryALTinterwordspacing}{\spaceskip=\fontdimen2\font plus
\BIBentryALTinterwordstretchfactor\fontdimen3\font minus
  \fontdimen4\font\relax}
\providecommand{\BIBforeignlanguage}[2]{{%
\expandafter\ifx\csname l@#1\endcsname\relax
\typeout{** WARNING: IEEEtran.bst: No hyphenation pattern has been}%
\typeout{** loaded for the language `#1'. Using the pattern for}%
\typeout{** the default language instead.}%
\else
\language=\csname l@#1\endcsname
\fi
#2}}
\providecommand{\BIBdecl}{\relax}
\BIBdecl

\bibitem{ronneberger2015u}
O.~Ronneberger, P.~Fischer, and T.~Brox, ``U-net: Convolutional networks for
  biomedical image segmentation,'' in \emph{International Conference on Medical
  image computing and computer-assisted intervention}.\hskip 1em plus 0.5em
  minus 0.4em\relax Springer, 2015, pp. 234--241.

\bibitem{yuan2021multi}
L.~Yuan, Y.~Li, Y.~Si, J.~Ren, Y.~Yang, Y.~Gong, Y.~Xia, Z.~Tong, and L.~Tong,
  ``Multi-objects change detection based on res-unet,'' in \emph{2021 IEEE
  International Geoscience and Remote Sensing Symposium IGARSS}.\hskip 1em plus
  0.5em minus 0.4em\relax IEEE, 2021, pp. 4364--4367.

\bibitem{jung2021boundary}
H.~Jung, H.-S. Choi, and M.~Kang, ``Boundary enhancement semantic segmentation
  for building extraction from remote sensed image,'' \emph{IEEE Transactions
  on Geoscience and Remote Sensing}, vol.~60, pp. 1--12, 2021.

\bibitem{li2021multistage}
R.~Li, S.~Zheng, C.~Duan, J.~Su, and C.~Zhang, ``Multistage attention resu-net
  for semantic segmentation of fine-resolution remote sensing images,''
  \emph{IEEE Geoscience and Remote Sensing Letters}, vol.~19, pp. 1--5, 2021.

\bibitem{ghosh2018stacked}
A.~Ghosh, M.~Ehrlich, S.~Shah, L.~S. Davis, and R.~Chellappa, ``Stacked u-nets
  for ground material segmentation in remote sensing imagery,'' in
  \emph{Proceedings of the IEEE Conference on Computer Vision and Pattern
  Recognition Workshops}, 2018, pp. 257--261.

\bibitem{chen2017rethinking}
L.-C. Chen, G.~Papandreou, F.~Schroff, and H.~Adam, ``Rethinking atrous
  convolution for semantic image segmentation,'' \emph{arXiv preprint
  arXiv:1706.05587}, 2017.

\bibitem{chen2018encoder}
L.-C. Chen, Y.~Zhu, G.~Papandreou, F.~Schroff, and H.~Adam, ``Encoder-decoder
  with atrous separable convolution for semantic image segmentation,'' in
  \emph{Proceedings of the European conference on computer vision (ECCV)},
  2018, pp. 801--818.

\bibitem{hamaguchi2018effective}
R.~Hamaguchi, A.~Fujita, K.~Nemoto, T.~Imaizumi, and S.~Hikosaka, ``Effective
  use of dilated convolutions for segmenting small object instances in remote
  sensing imagery,'' in \emph{2018 IEEE winter conference on applications of
  computer vision (WACV)}.\hskip 1em plus 0.5em minus 0.4em\relax IEEE, 2018,
  pp. 1442--1450.

\bibitem{nogueira2019dynamic}
K.~Nogueira, M.~Dalla~Mura, J.~Chanussot, W.~R. Schwartz, and J.~A. Dos~Santos,
  ``Dynamic multicontext segmentation of remote sensing images based on
  convolutional networks,'' \emph{IEEE Transactions on Geoscience and Remote
  Sensing}, vol.~57, no.~10, pp. 7503--7520, 2019.

\bibitem{marmanis2018classification}
D.~Marmanis, K.~Schindler, J.~D. Wegner, S.~Galliani, M.~Datcu, and U.~Stilla,
  ``Classification with an edge: Improving semantic image segmentation with
  boundary detection,'' \emph{ISPRS Journal of Photogrammetry and Remote
  Sensing}, vol. 135, pp. 158--172, 2018.

\bibitem{wang2017gated}
H.~Wang, Y.~Wang, Q.~Zhang, S.~Xiang, and C.~Pan, ``Gated convolutional neural
  network for semantic segmentation in high-resolution images,'' \emph{Remote
  Sensing}, vol.~9, no.~5, p. 446, 2017.

\bibitem{mou2020relation}
L.~Mou, Y.~Hua, and X.~X. Zhu, ``Relation matters: Relational context-aware
  fully convolutional network for semantic segmentation of high-resolution
  aerial images,'' \emph{IEEE Transactions on Geoscience and Remote Sensing},
  vol.~58, no.~11, pp. 7557--7569, 2020.

\bibitem{zhou2020class}
F.~Zhou, R.~Hang, and Q.~Liu, ``Class-guided feature decoupling network for
  airborne image segmentation,'' \emph{IEEE Transactions on Geoscience and
  Remote Sensing}, vol.~59, no.~3, pp. 2245--2255, 2020.

\bibitem{murez2020atlas}
Z.~Murez, T.~v. As, J.~Bartolozzi, A.~Sinha, V.~Badrinarayanan, and
  A.~Rabinovich, ``Atlas: End-to-end 3d scene reconstruction from posed
  images,'' in \emph{European conference on computer vision}.\hskip 1em plus
  0.5em minus 0.4em\relax Springer, 2020, pp. 414--431.

\bibitem{sun2021neuralrecon}
J.~Sun, Y.~Xie, L.~Chen, X.~Zhou, and H.~Bao, ``Neuralrecon: Real-time coherent
  3d reconstruction from monocular video,'' in \emph{Proceedings of the
  IEEE/CVF Conference on Computer Vision and Pattern Recognition}, 2021, pp.
  15\,598--15\,607.

\bibitem{qi20223d}
Z.~Qi, Z.~Zou, H.~Chen, and Z.~Shi, ``3d reconstruction of remote sensing
  mountain areas with tsdf-based neural networks,'' \emph{Remote Sensing},
  vol.~14, no.~17, p. 4333, 2022.

\bibitem{mildenhall2021nerf}
B.~Mildenhall, P.~P. Srinivasan, M.~Tancik, J.~T. Barron, R.~Ramamoorthi, and
  R.~Ng, ``Nerf: Representing scenes as neural radiance fields for view
  synthesis,'' \emph{Communications of the ACM}, vol.~65, no.~1, pp. 99--106,
  2021.

\bibitem{fu2022panoptic}
X.~Fu, S.~Zhang, T.~Chen, Y.~Lu, L.~Zhu, X.~Zhou, A.~Geiger, and Y.~Liao,
  ``Panoptic nerf: 3d-to-2d label transfer for panoptic urban scene
  segmentation,'' \emph{arXiv preprint arXiv:2203.15224}, 2022.

\bibitem{vaswani2017attention}
A.~Vaswani, N.~Shazeer, N.~Parmar, J.~Uszkoreit, L.~Jones, A.~N. Gomez,
  {\L}.~Kaiser, and I.~Polosukhin, ``Attention is all you need,''
  \emph{Advances in neural information processing systems}, vol.~30, 2017.

\bibitem{devlin2018bert}
J.~Devlin, M.-W. Chang, K.~Lee, and K.~Toutanova, ``Bert: Pre-training of deep
  bidirectional transformers for language understanding,'' \emph{arXiv preprint
  arXiv:1810.04805}, 2018.

\bibitem{brown2020language}
T.~Brown, B.~Mann, N.~Ryder, M.~Subbiah, J.~D. Kaplan, P.~Dhariwal,
  A.~Neelakantan, P.~Shyam, G.~Sastry, A.~Askell \emph{et~al.}, ``Language
  models are few-shot learners,'' \emph{Advances in neural information
  processing systems}, vol.~33, pp. 1877--1901, 2020.

\bibitem{fedus2021switch}
W.~Fedus, B.~Zoph, and N.~Shazeer, ``Switch transformers: Scaling to trillion
  parameter models with simple and efficient sparsity.''

\bibitem{dosovitskiy2020image}
A.~Dosovitskiy, L.~Beyer, A.~Kolesnikov, D.~Weissenborn, X.~Zhai,
  T.~Unterthiner, M.~Dehghani, M.~Minderer, G.~Heigold, S.~Gelly \emph{et~al.},
  ``An image is worth 16x16 words: Transformers for image recognition at
  scale,'' \emph{arXiv preprint arXiv:2010.11929}, 2020.

\bibitem{carion2020end}
N.~Carion, F.~Massa, G.~Synnaeve, N.~Usunier, A.~Kirillov, and S.~Zagoruyko,
  ``End-to-end object detection with transformers,'' in \emph{European
  conference on computer vision}.\hskip 1em plus 0.5em minus 0.4em\relax
  Springer, 2020, pp. 213--229.

\bibitem{radford2021learning}
A.~Radford, J.~W. Kim, C.~Hallacy, A.~Ramesh, G.~Goh, S.~Agarwal, G.~Sastry,
  A.~Askell, P.~Mishkin, J.~Clark \emph{et~al.}, ``Learning transferable visual
  models from natural language supervision,'' in \emph{International Conference
  on Machine Learning}.\hskip 1em plus 0.5em minus 0.4em\relax PMLR, 2021, pp.
  8748--8763.

\bibitem{hu2021unit}
R.~Hu and A.~Singh, ``Unit: Multimodal multitask learning with a unified
  transformer,'' in \emph{Proceedings of the IEEE/CVF International Conference
  on Computer Vision}, 2021, pp. 1439--1449.

\bibitem{kim2021vilt}
W.~Kim, B.~Son, and I.~Kim, ``Vilt: Vision-and-language transformer without
  convolution or region supervision,'' in \emph{International Conference on
  Machine Learning}.\hskip 1em plus 0.5em minus 0.4em\relax PMLR, 2021, pp.
  5583--5594.

\bibitem{chen2021remote}
H.~Chen, Z.~Qi, and Z.~Shi, ``Remote sensing image change detection with
  transformers,'' \emph{IEEE Transactions on Geoscience and Remote Sensing},
  vol.~60, pp. 1--14, 2021.

\bibitem{zheng2021rethinking}
S.~Zheng, J.~Lu, H.~Zhao, X.~Zhu, Z.~Luo, Y.~Wang, Y.~Fu, J.~Feng, T.~Xiang,
  P.~H. Torr \emph{et~al.}, ``Rethinking semantic segmentation from a
  sequence-to-sequence perspective with transformers,'' in \emph{Proceedings of
  the IEEE/CVF conference on computer vision and pattern recognition}, 2021,
  pp. 6881--6890.

\bibitem{schonberger2016structure}
J.~L. Schonberger and J.-M. Frahm, ``Structure-from-motion revisited,'' in
  \emph{Proceedings of the IEEE conference on computer vision and pattern
  recognition}, 2016, pp. 4104--4113.

\bibitem{zhang2020nerf++}
K.~Zhang, G.~Riegler, N.~Snavely, and V.~Koltun, ``Nerf++: Analyzing and
  improving neural radiance fields,'' \emph{arXiv preprint arXiv:2010.07492},
  2020.

\bibitem{barron2022mip}
J.~T. Barron, B.~Mildenhall, D.~Verbin, P.~P. Srinivasan, and P.~Hedman,
  ``Mip-nerf 360: Unbounded anti-aliased neural radiance fields,'' in
  \emph{Proceedings of the IEEE/CVF Conference on Computer Vision and Pattern
  Recognition}, 2022, pp. 5470--5479.

\bibitem{al2019land}
H.~A. Al-Najjar, B.~Kalantar, B.~Pradhan, V.~Saeidi, A.~A. Halin, N.~Ueda, and
  S.~Mansor, ``Land cover classification from fused dsm and uav images using
  convolutional neural networks,'' \emph{Remote Sensing}, vol.~11, no.~12, p.
  1461, 2019.

\bibitem{li2020deep}
W.~Li, Z.~Zou, and Z.~Shi, ``Deep matting for cloud detection in remote sensing
  images,'' \emph{IEEE Transactions on Geoscience and Remote Sensing}, vol.~58,
  no.~12, pp. 8490--8502, 2020.

\bibitem{barron2021mip}
J.~T. Barron, B.~Mildenhall, M.~Tancik, P.~Hedman, R.~Martin-Brualla, and P.~P.
  Srinivasan, ``Mip-nerf: A multiscale representation for anti-aliasing neural
  radiance fields,'' in \emph{Proceedings of the IEEE/CVF International
  Conference on Computer Vision}, 2021, pp. 5855--5864.

\bibitem{pan2019deep}
J.~Pan, X.~Han, W.~Chen, J.~Tang, and K.~Jia, ``Deep mesh reconstruction from
  single rgb images via topology modification networks,'' in \emph{Proceedings
  of the IEEE/CVF International Conference on Computer Vision}, 2019, pp.
  9964--9973.

\bibitem{kanazawa2018learning}
A.~Kanazawa, S.~Tulsiani, A.~A. Efros, and J.~Malik, ``Learning
  category-specific mesh reconstruction from image collections,'' in
  \emph{Proceedings of the European Conference on Computer Vision (ECCV)},
  2018, pp. 371--386.

\bibitem{poullis2009automatic}
C.~Poullis and S.~You, ``Automatic reconstruction of cities from remote sensor
  data,'' in \emph{2009 IEEE conference on computer vision and pattern
  recognition}.\hskip 1em plus 0.5em minus 0.4em\relax IEEE, 2009, pp.
  2775--2782.

\bibitem{zhi2021place}
S.~Zhi, T.~Laidlow, S.~Leutenegger, and A.~J. Davison, ``In-place scene
  labelling and understanding with implicit scene representation,'' in
  \emph{Proceedings of the IEEE/CVF International Conference on Computer
  Vision}, 2021, pp. 15\,838--15\,847.

\bibitem{kundu2022panoptic}
A.~Kundu, K.~Genova, X.~Yin, A.~Fathi, C.~Pantofaru, L.~J. Guibas,
  A.~Tagliasacchi, F.~Dellaert, and T.~Funkhouser, ``Panoptic neural fields: A
  semantic object-aware neural scene representation,'' in \emph{Proceedings of
  the IEEE/CVF Conference on Computer Vision and Pattern Recognition}, 2022,
  pp. 12\,871--12\,881.

\bibitem{dosovitskiy2017carla}
A.~Dosovitskiy, G.~Ros, F.~Codevilla, A.~Lopez, and V.~Koltun, ``Carla: An open
  urban driving simulator,'' in \emph{Conference on robot learning}.\hskip 1em
  plus 0.5em minus 0.4em\relax PMLR, 2017, pp. 1--16.

\bibitem{fu2019dual}
J.~Fu, J.~Liu, H.~Tian, Y.~Li, Y.~Bao, Z.~Fang, and H.~Lu, ``Dual attention
  network for scene segmentation,'' in \emph{Proceedings of the IEEE/CVF
  conference on computer vision and pattern recognition}, 2019, pp. 3146--3154.

\bibitem{9852475}
Y.~Wu, Z.~Zou, and Z.~Shi, ``Remote sensing novel view synthesis with implicit
  multiplane representations,'' \emph{IEEE Transactions on Geoscience and
  Remote Sensing}, vol.~60, pp. 1--13, 2022.

\bibitem{eteke2022semantic}
C.~Eteke, J.~Zhang, and E.~Steinbach, ``Semantic-srf: Sparse multi-view indoor
  semantic segmentation with stereo neural radiance fields,'' in \emph{2022
  10th European Workshop on Visual Information Processing (EUVIP)}.\hskip 1em
  plus 0.5em minus 0.4em\relax IEEE, 2022, pp. 1--6.

\bibitem{tseng2022cla}
W.-C. Tseng, H.-J. Liao, L.~Yen-Chen, and M.~Sun, ``Cla-nerf: Category-level
  articulated neural radiance field,'' \emph{arXiv preprint arXiv:2202.00181},
  2022.

\bibitem{vora2021nesf}
S.~Vora, N.~Radwan, K.~Greff, H.~Meyer, K.~Genova, M.~S. Sajjadi, E.~Pot,
  A.~Tagliasacchi, and D.~Duckworth, ``Nesf: Neural semantic fields for
  generalizable semantic segmentation of 3d scenes,'' \emph{arXiv preprint
  arXiv:2111.13260}, 2021.

\bibitem{tschernezki2022neural}
V.~Tschernezki, I.~Laina, D.~Larlus, and A.~Vedaldi, ``Neural feature fusion
  fields: 3d distillation of self-supervised 2d image representations,''
  \emph{arXiv preprint arXiv:2209.03494}, 2022.

\bibitem{yuan2022nerf}
Y.-J. Yuan, Y.-T. Sun, Y.-K. Lai, Y.~Ma, R.~Jia, and L.~Gao, ``Nerf-editing:
  geometry editing of neural radiance fields,'' in \emph{Proceedings of the
  IEEE/CVF Conference on Computer Vision and Pattern Recognition}, 2022, pp.
  18\,353--18\,364.

\bibitem{badrinarayanan2017segnet}
V.~Badrinarayanan, A.~Kendall, and R.~Cipolla, ``Segnet: A deep convolutional
  encoder-decoder architecture for image segmentation,'' \emph{IEEE
  transactions on pattern analysis and machine intelligence}, vol.~39, no.~12,
  pp. 2481--2495, 2017.

\bibitem{qi2022remote}
Z.~Qi, Z.~Zou, H.~Chen, and Z.~Shi, ``Remote sensing image segmentation based
  on implicit 3d scene representation,'' \emph{IEEE Geoscience and Remote
  Sensing Letters}, 2022.

\bibitem{song2022lslpct}
Y.~Song, F.~He, Y.~Duan, T.~Si, and J.~Bai, ``Lslpct: An enhanced local
  semantic learning transformer for 3-d point cloud analysis,'' \emph{IEEE
  Transactions on Geoscience and Remote Sensing}, vol.~60, pp. 1--13, 2022.

\bibitem{ye2021shelf}
Y.~Ye, S.~Tulsiani, and A.~Gupta, ``Shelf-supervised mesh prediction in the
  wild,'' in \emph{Proceedings of the IEEE/CVF Conference on Computer Vision
  and Pattern Recognition}, 2021, pp. 8843--8852.

\bibitem{wang2018understanding}
P.~Wang, P.~Chen, Y.~Yuan, D.~Liu, Z.~Huang, X.~Hou, and G.~Cottrell,
  ``Understanding convolution for semantic segmentation,'' in \emph{2018 IEEE
  winter conference on applications of computer vision (WACV)}.\hskip 1em plus
  0.5em minus 0.4em\relax Ieee, 2018, pp. 1451--1460.

\bibitem{liu2021swin}
Z.~Liu, Y.~Lin, Y.~Cao, H.~Hu, Y.~Wei, Z.~Zhang, S.~Lin, and B.~Guo, ``Swin
  transformer: Hierarchical vision transformer using shifted windows,'' in
  \emph{Proceedings of the IEEE/CVF International Conference on Computer
  Vision}, 2021, pp. 10\,012--10\,022.

\bibitem{hu2021vmnet}
Z.~Hu, X.~Bai, J.~Shang, R.~Zhang, J.~Dong, X.~Wang, G.~Sun, H.~Fu, and C.-L.
  Tai, ``Vmnet: Voxel-mesh network for geodesic-aware 3d semantic
  segmentation,'' in \emph{Proceedings of the IEEE/CVF International Conference
  on Computer Vision}, 2021, pp. 15\,488--15\,498.

\end{thebibliography}

\vfill



\end{document}